\documentclass[11pt]{article}

\usepackage[final]{acl}

\usepackage{times}
\usepackage{latexsym}

\usepackage[T1]{fontenc}

\usepackage[utf8]{inputenc}

\usepackage{microtype}

\usepackage{inconsolata}

\usepackage{graphicx}
\usepackage[table]{xcolor}
\usepackage{tikz}
\usepackage{pgfplots}
\pgfplotsset{compat=1.18}

\usepackage{booktabs}
\usepackage{amsmath,amssymb}
\usepackage{tcolorbox}
\usepackage{multirow}
\usepackage{subcaption}
\usepackage{wrapfig}
\usepackage{algorithm}
\usepackage{algpseudocode}
\usepackage{listings}

\lstdefinestyle{pythonstyle}{
    language=Python,
    basicstyle=\small\ttfamily,
    keywordstyle=\color{mypurple}\bfseries,
    commentstyle=\color{mygray}\itshape,
    stringstyle=\color{myorange},
    numbers=none,
    showstringspaces=false,
    breaklines=true,
    frame=single,
    rulecolor=\color{gray!40},
    xleftmargin=1em,
    xrightmargin=1em,
    tabsize=4,
}

\tcbset{
  promptbox/.style={
    width=\textwidth,
    colback=gray!8,
    colframe=gray!40,
    fontupper=\small\ttfamily,
    boxrule=0.4pt,
    arc=2pt,
    left=4pt, right=4pt, top=4pt, bottom=4pt,
  }
}
\definecolor{mylightorange}{HTML}{fdb863}
\definecolor{mylightpurple}{HTML}{B2ABD2}
\definecolor{mypurple}{HTML}{5e3c99}
\definecolor{myorange}{HTML}{e66101}
\definecolor{mygray}{HTML}{888780}
\definecolor{mygreen}{HTML}{009E73}
\definecolor{mylightgreen}{HTML}{95DFC8}

\newcommand{\CTFAlign}{\mbox{CTFAlign}}
\newcommand{\MDPAlign}{\mbox{MDPAlign}}
\newcommand{\SimAlign}{\mbox{SimAlign}}

\title{Scaling Unsupervised Word Alignment to Documents\\ via Structural Constraints}

\author{Michelle Wastl\quad Jannis Vamvas\quad Rico Sennrich
\vspace{0.1cm}\\
 Department of Computational Linguistics, University of Zurich\\
\texttt{\{michelle.wastl,jannisnikos.vamvas,rico.sennrich\}@uzh.ch}
}

\begin{document}
\maketitle

\begin{abstract}
Word alignment has traditionally been studied between sentences, but many cross-lingual tasks increasingly require correspondences across full documents. While recent multilingual embedding models can encode long inputs, we show that applying algorithms designed for sentences directly to documents leads to performance degradation.
To address this, we introduce \textbf{\CTFAlign{}}, a lightweight, training-free approach for document-level word alignment. \CTFAlign{} applies a \textbf{c}oarse-\textbf{t}o-\textbf{f}ine refinement strategy that restricts the alignment search space to semantically similar regions.
Additionally, we introduce \textbf{\MDPAlign{}}, a simpler alternative that constrains alignments by position with a \textbf{m}ain \textbf{d}iagonal \textbf{p}rior. Both approaches operate directly on full documents without relying on sentence segmentation or sentence alignment. We evaluate these methods across six language pairs varying in typological distance, resourcedness, and document length. Averaged over three models, \CTFAlign{} reduces word alignment error rate from 0.412 to 0.326. 
These gains transfer downstream, leading to improvements in document-level translation coverage evaluation and recognition of semantic differences. 
We release \CTFAlign{} as a Python package and make the code and data to reproduce our experiments publicly available.\footnote{Package:~\url{https://github.com/ZurichNLP/CTFAlign};\\
experimental code:~\url{https://github.com/ZurichNLP/document-level-word-alignment};\\
data:~\url{https://huggingface.co/datasets/ZurichNLP/document-level-word-alignment}}

\end{abstract}

\section{Introduction}



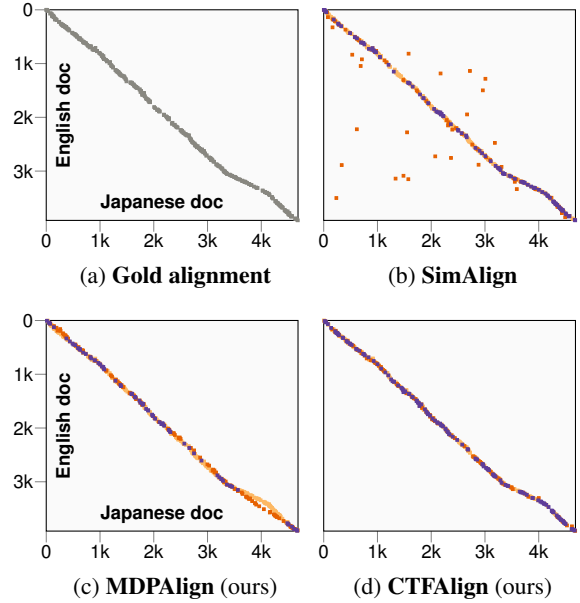
\begin{figure}[!t]
    \centering
    \makebox[\columnwidth][l]{%
    \hspace*{4mm}%
    \begin{tabular}{@{}c@{\hspace{0.5mm}}c@{}}
    \begin{subfigure}[t]{0.47\columnwidth}
        \centering
\begin{tikzpicture}[trim axis left, trim axis right]
\begin{axis}[
  width=0.92\linewidth,
  scale only axis,
  axis equal image,
  axis background/.style={fill=black!2},
  xmin=-0.5,
  xmax=4683.5,
  ymin=-0.5,
  ymax=3917.5,
  y dir=reverse,
  xtick={0,1000,2000,3000,4000},
  xticklabels={0,1k,2k,3k,4k},
  ytick={0,1000,2000,3000},
  yticklabels={0,1k,2k,3k},
  tick label style={
    font={\scriptsize\sffamily},
    inner sep=0.5pt,
  },
  tick align=outside,
  xtick pos=bottom,
  ytick pos=left,
  enlargelimits=false,
  clip=true,
]
  \addplot[
    only marks,
    mark=square*,
    mark size=0.5pt,
    color=mygray,
    fill=mygray,
  ] coordinates {
      (3328,3012) (3734,3218) (354,318) (2261,2015) (4434,3721) (3528,3125) (2646,2340) (4345,3634)
      (3073,2788) (1903,1685) (4637,3893) (1216,1039) (2444,2191) (1214,1046) (915,709) (2187,1951)
      (1291,1122) (2685,2406) (881,733) (628,548) (2845,2572) (4310,3603) (901,742) (3062,2769)
      (3904,3304) (2715,2460) (1802,1563) (4349,3637) (4028,3367) (2275,2024) (542,461) (123,113)
      (813,696) (3461,3092) (12,4) (4513,3799) (2861,2595) (2112,1893) (2816,2538) (556,472)
      (1676,1445) (1926,1709) (2068,1856) (1243,1075) (4105,3403) (831,680) (694,601) (2761,2489)
      (4679,3916) (4469,3755) (992,802) (2606,2326) (2521,2265) (1072,917) (2858,2588) (1814,1558)
      (2512,2242) (4068,3386) (825,683) (3326,3012) (2730,2469) (2104,1885) (2519,2261) (1607,1380)
      (1655,1428) (1633,1402) (281,270) (2280,2020) (4344,3634) (600,521) (777,671) (1130,957)
      (1308,1146) (335,304) (692,601) (3365,3054) (2382,2142) (2023,1827) (1870,1656) (3715,3210)
      (2384,2141) (4139,3420) (4502,3787) (3078,2785) (718,612) (2775,2529) (1039,848) (4437,3729)
      (2875,2613) (1673,1436) (2096,1889) (134,135) (2336,2116) (1062,864) (1923,1707) (3225,2911)
      (1481,1295) (2863,2587) (3814,3258) (537,460) (1288,1120) (1901,1688) (386,349) (638,551)
      (241,223) (1072,916) (1649,1421) (1057,887) (3364,3053) (824,682) (3189,2897) (1033,855)
      (317,294) (163,179) (1739,1481) (1627,1389) (1080,910) (4363,3656) (1835,1611) (526,455)
      (194,189) (3814,3257) (890,748) (2279,2020) (625,528) (1926,1707) (609,540) (2605,2322)
      (3030,2745) (3931,3314) (1593,1356) (4651,3888) (4269,3564) (345,310) (899,745) (1752,1527)
      (2273,2024) (3069,2792) (4295,3590) (1474,1288) (1816,1558) (497,438) (1378,1205) (1418,1228)
      (2950,2668) (2188,1952) (1061,865) (4024,3360) (1492,1331) (130,108) (4305,3599) (3586,3152)
      (2693,2423) (3076,2785) (3345,3036) (2189,1954) (1346,1172) (105,101) (4196,3484) (683,588)
      (2876,2614) (397,353) (2433,2185) (1177,995) (2087,1866) (4386,3677) (3052,2757) (2509,2243)
      (3092,2799) (1187,975) (2685,2407) (3352,3032) (3647,3177) (695,604) (15,6) (1701,1454)
      (2888,2606) (1392,1219) (2066,1858) (1921,1713) (4102,3405) (3275,2955) (269,257) (3497,3108)
      (3698,3201) (402,363) (1455,1255) (4067,3388) (547,480) (2257,2017) (1371,1203) (4073,3394)
      (4446,3738) (1,0) (340,307) (4511,3807) (2813,2540) (2693,2428) (4266,3562) (429,386)
      (3651,3185) (1638,1419) (731,619) (1129,959) (108,120) (3494,3106) (3686,3197) (32,20)
      (1856,1645) (121,115) (21,15) (861,727) (1674,1436) (1056,886) (1760,1516) (2616,2336)
      (2423,2176) (1596,1367) (873,738) (4338,3627) (3440,3083) (696,608) (164,179) (2390,2138)
      (4155,3446) (1036,851) (2244,1995) (1180,981) (2977,2714) (1037,848) (1361,1180) (2407,2168)
      (158,159) (375,333) (371,320) (1788,1573) (2277,2020) (2714,2451) (3162,2875) (362,330)
      (4511,3806) (4202,3490) (3917,3313) (3224,2914) (1563,1350) (1825,1597) (3246,2937) (73,71)
      (1207,1029) (1338,1157) (2342,2119) (2898,2599) (2936,2626) (3168,2876) (805,699) (2903,2633)
      (311,290) (369,321) (2349,2102) (1441,1238) (1321,1145) (2527,2256) (3109,2821) (3418,3072)
      (1128,951) (2553,2279) (1031,847) (4355,3648) (2070,1861) (419,370) (2667,2384) (1576,1353)
      (620,534) (2651,2337) (3517,3110) (2801,2557) (2595,2312) (3645,3183) (985,800) (862,738)
      (4381,3671) (4325,3619) (2902,2623) (2956,2677) (1075,915) (4364,3655) (4553,3838) (3814,3259)
      (324,302) (2612,2333) (2882,2608) (1366,1193) (1458,1262) (763,653) (577,495) (743,630)
      (1764,1512) (1920,1709) (3325,3012) (82,63) (885,728) (3410,3065) (2523,2257) (4099,3409)
      (4491,3749) (1897,1670) (3773,3241) (733,620) (3182,2883) (1920,1707) (2270,2030) (3863,3288)
      (1684,1452) (3068,2792) (236,225) (4142,3432) (1779,1543) (4573,3835) (3455,3088) (1785,1581)
      (1865,1661) (946,764) (1763,1512) (4201,3488) (3267,2951) (3122,2828) (1340,1157) (942,765)
      (4459,3746) (1317,1141) (3535,3132) (3778,3242) (4530,3817) (2770,2517) (4558,3845) (4559,3846)
      (1819,1546) (103,79) (2920,2650) (2737,2475) (2763,2509) (312,289) (2243,1998) (1366,1195)
      (3273,2956) (2548,2284) (4296,3590) (572,501) (1932,1719) (355,319) (2584,2316) (142,126)
      (2841,2574) (1408,1224) (1310,1142) (4218,3512) (3204,2907) (4640,3893) (3295,2988) (4630,3879)
      (292,281) (804,677) (659,565) (3812,3259) (1793,1578) (2528,2256) (3688,3197) (4401,3680)
      (2007,1809) (171,172) (4,0) (1578,1364) (4667,3905) (2229,1981) (575,495) (4505,3796)
      (2282,2040) (2652,2337) (3558,3145) (3325,3011) (3304,3005) (1443,1237) (1111,931) (2066,1857)
      (2497,2238) (2896,2601) (480,426) (315,291) (4383,3673) (3697,3198) (1237,1060) (4523,3781)
      (699,610) (1341,1157) (3045,2761) (3606,3160) (303,287) (4263,3558) (3318,3021) (4204,3493)
      (404,363) (4299,3594) (174,171) (287,266) (1184,977) (2786,2525) (945,771) (2985,2706)
      (1736,1482) (3306,3001) (4479,3765) (1007,821) (520,450) (4270,3565) (2909,2628) (2558,2296)
  };
  \node[anchor=center, font={\scriptsize\sffamily\bfseries}, rotate=90] at (axis cs:350.8,1958.5) {English doc};
  \node[anchor=south, inner sep=1.5pt, font={\scriptsize\sffamily\bfseries}] at (axis cs:2154.14,3839.14) {Japanese doc};
\end{axis}
\end{tikzpicture}
        \caption{\textbf{Gold alignment}}
        \label{fig:gold}
    \end{subfigure} &
    \begin{subfigure}[t]{0.47\columnwidth}
        \centering
\begin{tikzpicture}[trim axis left, trim axis right]
\begin{axis}[
  width=0.92\linewidth,
  scale only axis,
  axis equal image,
  axis background/.style={fill=black!2},
  xmin=-0.5,
  xmax=4683.5,
  ymin=-0.5,
  ymax=3917.5,
  y dir=reverse,
  xtick={0,1000,2000,3000,4000},
  xticklabels={0,1k,2k,3k,4k},
  ytick=\empty,
  yticklabels={},
  ymajorticks=false,
  tick label style={
    font={\scriptsize\sffamily},
    inner sep=0.5pt,
  },
  tick align=outside,
  xtick pos=bottom,
  ytick pos=left,
  enlargelimits=false,
  clip=true,
]
  \addplot[
    only marks,
    mark=square*,
    mark size=0.5pt,
    color=mylightorange,
    fill=mylightorange,
  ] coordinates {
      (769,646) (177,164) (1964,1753) (134,135) (1054,891) (2324,2077) (3465,3096) (4141,3429)
      (3778,3242) (95,88) (1834,1612) (2598,2312) (3070,2787) (4677,3904) (4193,3483) (2783,2525)
      (2693,2430) (471,433) (338,303) (2983,2709) (2302,2062) (1503,1329) (171,172) (3714,3213)
      (2877,2613) (582,506) (469,433) (1907,1677) (2012,1806) (601,521) (2678,2394) (220,199)
      (962,785) (2876,2614) (3535,3132) (1674,1436) (230,225) (0,1) (2121,1899) (1925,1707)
      (992,802) (3060,2774) (4210,3502) (888,748) (1014,825) (1060,887) (1000,817) (3649,3184)
      (4023,3360) (1416,1228) (3525,3127) (1880,1649) (147,153) (2725,2457) (3531,3134) (4009,3354)
      (2283,2040) (283,267) (1354,1164) (1397,1218) (3614,3169) (1632,1403) (1881,1649) (1061,865)
      (1399,1211) (1958,1762) (1337,1160) (1026,842) (2396,2156) (2625,2343) (2014,1833) (3156,2865)
      (910,714) (2224,1975) (3154,2865) (1016,837) (3256,2945) (40,32) (4102,3405) (2243,1998)
      (4377,3667) (1047,876) (923,751) (109,119) (185,161) (1836,1612) (2833,2580) (33,21)
      (237,225) (301,287) (665,577) (3203,2906) (3440,3083) (892,748) (959,780) (3961,3326)
      (2161,1915) (178,165) (2380,2144) (2510,2242) (3185,2881) (4105,3403) (1893,1672) (4335,3625)
      (1438,1248) (907,714) (2018,1836) (1322,1111) (2997,2698) (2271,2058) (442,401) (798,676)
      (2352,2096) (3166,2870) (3810,3260) (585,506) (2471,2206) (1441,1237) (2677,2394) (2176,1931)
      (1340,1157) (2279,2020) (1396,1218) (3701,3200) (2714,2451) (2283,2038) (3244,2940) (427,382)
      (1470,1283) (1603,1371) (2938,2661) (2077,1877) (3040,2752) (136,136) (3369,3056) (3441,3083)
      (2693,2422) (1373,1204) (1609,1376) (246,237) (3433,3083) (1840,1610) (2246,1999) (3537,3134)
      (635,556) (2990,2701) (315,291) (879,736) (4352,3642) (2256,2008) (179,168) (2396,2157)
      (4513,3799) (2261,2015) (2451,2214) (627,545) (4351,3641) (2612,2333) (2653,2337) (1025,831)
      (1065,863) (1921,1712) (2696,2417) (2328,2107) (3297,2987) (1648,1421) (2805,2553) (706,582)
      (2067,1858) (1933,1717) (885,729) (3561,3140) (1996,1796) (2685,2407) (3565,3151) (2936,2625)
      (343,314) (1180,980) (1484,1302) (1053,891) (4443,3736) (1830,1625) (1733,1486) (2164,1917)
      (3609,3162) (275,247) (1713,1459) (3059,2781) (2470,2206) (208,210) (4067,3386) (2327,2104)
      (1565,1350) (2113,1893) (4473,3762) (1236,1060) (1934,1725) (811,696) (825,683) (676,570)
      (3455,3086) (1997,1795) (1028,846) (194,189) (2474,2204) (3221,2919) (2285,2046) (2641,2345)
      (461,418) (3670,3192) (4331,3622) (4294,3591) (4031,3368) (458,422) (2421,2176) (3049,2758)
      (4467,3755) (3080,2785) (906,716) (1064,863) (953,779) (1220,1056) (1029,844) (1352,1167)
      (163,179) (2915,2655) (806,700) (2763,2509) (3521,3119) (4577,3835) (1668,1443) (4368,3659)
      (1901,1689) (4073,3394) (720,612) (2347,2101) (1507,1326) (1005,812) (1922,1713) (3552,3138)
      (2810,2542) (1896,1669) (2991,2697) (4197,3485) (863,738) (2129,1897) (4624,3879) (4639,3893)
      (3099,2800) (1773,1537) (1383,1198) (2150,1913) (1929,1724) (2244,1998) (2832,2582) (1777,1541)
      (2348,2102) (2317,2087) (588,518) (4068,3388) (965,784) (38,29) (3260,2944) (3308,3017)
      (3609,3161) (4282,3575) (2871,2615) (652,558) (4213,3508) (3709,3209) (4234,3527) (1217,1035)
      (3309,3018) (3686,3197) (3363,3053) (2886,2604) (670,573) (1542,1341) (4115,3414) (4509,3801)
      (3457,3089) (1041,860) (2128,1897) (520,450) (213,212) (74,72) (693,601) (2019,1829)
      (4418,3713) (2704,2447) (1843,1601) (648,559) (3982,3342) (2574,2303) (2086,1865) (3259,2950)
      (4212,3506) (1475,1288) (3870,3295) (2746,2473) (1349,1170) (3116,2833) (3060,2781) (459,422)
      (2846,2572) (1042,858)
  };
  \addplot[
    only marks,
    mark=square*,
    mark size=0.5pt,
    color=myorange,
    fill=myorange,
  ] coordinates {
      (169,178) (4373,3661) (1131,940) (3308,2922) (2571,2301) (4237,3496) (4414,3708) (2174,1184)
      (3184,2475) (4492,3858) (4431,3782) (1551,2279) (2960,1498) (78,137) (3936,3336) (977,765)
      (2462,2201) (1164,975) (2765,2505) (3666,3197) (164,326) (2313,2151) (578,501) (4348,3665)
      (3960,3340) (2667,1946) (1485,1306) (2881,2882) (2859,2591) (281,275) (3021,2744) (4283,3571)
      (657,565) (1747,1519) (335,2888) (3096,2795) (3535,3133) (4368,3667) (2392,2229) (2937,2661)
      (1730,1499) (688,1045) (3008,1283) (1484,3087) (2321,2093) (710,921) (1952,1745) (235,3501)
      (531,835) (2864,2596) (488,432) (2075,2726) (616,2222) (3555,3199) (2574,2306) (1337,3140)
      (621,535) (2368,2755) (730,616) (1909,1697) (1579,816) (2289,2051) (627,539) (955,785)
      (4407,3696) (404,368) (4437,3730) (961,784) (3592,3335) (3585,3152) (2720,1137) (63,50)
      (211,213) (1574,3151)
  };
  \addplot[
    only marks,
    mark=square*,
    mark size=0.5pt,
    color=mypurple,
    fill=mypurple,
  ] coordinates {
      (4396,3689) (840,726) (4315,3607) (3290,2983) (3267,2952) (757,643) (27,16) (2223,1974)
      (2170,1908) (4433,3719) (2901,2622) (2495,2232) (2095,1890) (48,40) (4417,3712) (3338,3040)
      (4500,3785) (3243,2940) (698,606) (579,508) (2823,2568) (4126,3418) (2117,1905) (3485,3102)
      (3675,3197) (1206,1022) (2175,1933) (2648,2339) (1854,1643) (4346,3631) (3361,3056) (2565,2295)
      (3735,3218) (1046,878) (967,787) (3931,3314) (3335,3045) (4438,3730) (4308,3601) (2165,1917)
      (3557,3144) (3365,3054) (4162,3454) (2505,2249) (3909,3313) (908,712) (4340,3628) (2661,2371)
      (971,791) (4208,3497) (1303,1150) (3270,2959) (2228,1980) (1656,1429) (3198,2904) (4450,3726)
      (901,742) (3737,3216) (3797,3255) (4278,3570) (4511,3806) (4268,3563) (2054,1844) (3905,3307)
      (4448,3739) (4378,3670) (2196,1938) (2948,2670) (4573,3835) (14,6) (699,610) (4312,3605)
      (3007,2721) (3858,3287) (2143,1910) (4161,3453) (325,304) (268,256) (4003,3348) (739,617)
      (2366,2128) (1858,1646) (3107,2816) (1583,1361) (929,752) (1598,1371) (3455,3087) (187,181)
      (2183,1945) (278,273) (46,37) (3716,3210) (4670,3908) (21,15) (1046,879) (943,770)
      (618,531) (4284,3577) (1792,1575) (4558,3845) (4065,3390) (491,434) (4679,3916) (1569,1354)
      (2185,1947) (3699,3201) (4318,3614) (3,1) (4175,3466) (1944,1736)
  };
\end{axis}
\end{tikzpicture}
        \caption{\textbf{SimAlign}}
        \label{fig:simalign}
    \end{subfigure} \\[7mm]
    \begin{subfigure}[t]{0.47\columnwidth}
        \centering
\begin{tikzpicture}[trim axis left, trim axis right]
\begin{axis}[
  width=0.92\linewidth,
  scale only axis,
  axis equal image,
  axis background/.style={fill=black!2},
  xmin=-0.5,
  xmax=4683.5,
  ymin=-0.5,
  ymax=3917.5,
  y dir=reverse,
  xtick={0,1000,2000,3000,4000},
  xticklabels={0,1k,2k,3k,4k},
  ytick={0,1000,2000,3000},
  yticklabels={0,1k,2k,3k},
  tick label style={
    font={\scriptsize\sffamily},
    inner sep=0.5pt,
  },
  tick align=outside,
  xtick pos=bottom,
  ytick pos=left,
  enlargelimits=false,
  clip=true,
]
  \addplot[
    only marks,
    mark=square*,
    mark size=0.5pt,
    color=mylightorange,
    fill=mylightorange,
  ] coordinates {
      (1839,1610) (1286,1122) (2080,1875) (3430,3077) (2690,2410) (1602,1371) (3154,2865) (947,758)
      (695,604) (4544,3820) (104,99) (3012,2735) (2780,2525) (1836,1611) (2055,1843) (3191,2879)
      (3468,3097) (158,159) (4553,3838) (649,559) (2282,2040) (811,696) (992,802) (2813,2540)
      (380,348) (689,600) (4268,3563) (1141,935) (4334,3624) (4307,3600) (2767,2482) (3066,2795)
      (4277,3569) (3869,3295) (1181,980) (1282,1131) (4142,3422) (678,570) (4611,3882) (1291,1122)
      (4248,3542) (2210,1960) (1938,1743) (3296,2987) (3202,2902) (671,572) (3437,3081) (2761,2489)
      (538,459) (3309,3018) (185,161) (2883,2607) (3209,2923) (1990,1777) (2732,2452) (2562,2294)
      (1762,1514) (96,86) (1208,1026) (3030,2745) (3588,3157) (852,726) (3649,3184) (2745,2473)
      (4512,3800) (4157,3448) (2691,2414) (2964,2687) (1778,1541) (3932,3320) (396,356) (1441,1237)
      (3271,2956) (2323,2078) (3934,3315) (1056,886) (1815,1558) (4351,3638) (1633,1402) (2208,1960)
      (315,291) (1130,954) (216,203) (1807,1548) (4103,3403) (4197,3485) (1381,1199) (1921,1709)
      (2511,2243) (3703,3202) (4079,3400) (3647,3177) (3624,3172) (1231,1070) (1442,1238) (1925,1709)
      (3737,3215) (845,726) (1738,1482) (116,109) (1958,1762) (3112,2827) (2917,2655) (312,289)
      (4139,3420) (3444,3083) (802,677) (151,143) (2716,2458) (3412,3068) (320,298) (145,155)
      (1337,1158) (249,229) (96,85) (1231,1069) (3609,3161) (1844,1605) (343,314) (4667,3905)
      (320,300) (1920,1712) (3677,3194) (2410,2166) (690,601) (1172,1000) (2536,2271) (2476,2217)
      (733,620) (1990,1776) (1749,1530) (3566,3148) (1835,1612) (4099,3409) (2330,2106) (2198,1935)
      (74,72) (2329,2111) (1460,1273) (2908,2636) (2936,2625) (4491,3749) (4150,3441) (3833,3268)
      (3940,3322) (2598,2312) (4245,3539) (230,225) (3060,2780) (610,539) (2328,2107) (2241,2004)
      (32,20) (2765,2500) (696,607) (3337,3041) (3673,3187) (2803,2554) (3907,3305) (786,665)
      (403,362) (1846,1619) (1312,1142) (4666,3905) (985,801) (1364,1183) (4471,3762) (894,748)
      (4179,3467) (4356,3649) (3812,3259) (3060,2773) (891,748) (4045,3372) (396,355) (2429,2182)
      (3230,2930) (512,447) (3348,3036) (2557,2298) (3988,3344) (776,670) (3329,3012) (3265,2942)
      (213,212) (1399,1211) (1911,1699) (4468,3755) (4159,3450) (4142,3423) (2115,1892) (178,165)
      (3388,3058) (204,198) (2766,2482) (444,397) (1991,1776) (1644,1414) (3737,3216) (800,675)
      (2573,2301) (1317,1141) (2788,2533) (1960,1760) (4149,3440) (4220,3514) (2321,2088) (1896,1670)
      (1482,1300) (4459,3746) (2721,2463) (2279,2020) (92,96) (1755,1524) (2874,2618) (1627,1387)
      (2977,2714) (2653,2337) (200,197) (4299,3594) (2936,2626) (3099,2800) (3008,2727) (1453,1257)
      (3833,3269) (2563,2294) (2282,2039) (4067,3387) (597,523) (2317,2086) (1920,1713) (2735,2478)
      (276,260) (136,136) (1509,1324) (4167,3460) (190,192) (1397,1218) (3797,3255) (3795,3254)
      (3153,2865) (3921,3313) (4313,3606) (1625,1390) (3819,3266) (4257,3550) (1134,948) (4259,3552)
      (2877,2614) (2827,2564) (2219,1971) (1382,1200) (1130,957) (573,503) (3078,2784) (1833,1614)
      (4470,3753) (1354,1164) (3186,2881) (3747,3226) (4063,3383) (2904,2631) (3808,3256) (847,726)
      (831,680) (1829,1626) (4198,3486) (4677,3904) (73,69) (3169,2877) (810,697) (1918,1710)
      (2037,1815) (1630,1398) (3387,3058) (1638,1419) (1824,1598) (4340,3628) (3243,2941) (2462,2202)
      (924,751) (25,2) (4359,3652) (740,616) (3074,2790) (238,225) (2513,2243) (627,545)
      (1921,1707) (73,71) (4068,3387) (3678,3194) (4640,3893) (3673,3190) (4323,3618) (1422,1228)
      (3304,3006) (1041,861) (365,324) (2989,2701) (2416,2173) (1548,1340) (963,784) (2878,2614)
      (923,751) (326,304) (3634,3176) (3503,3108) (912,708) (3143,2850) (549,478) (2771,2518)
      (3456,3087) (4393,3687) (2954,2676) (3274,2956) (4227,3521) (2453,2215) (1474,1287) (1754,1528)
      (1580,1363) (89,91) (2435,2188) (4322,3617) (4660,3900) (1564,1350) (1431,1233) (2922,2650)
      (2881,2608) (3019,2742) (1798,1563) (1924,1707) (1051,872) (3961,3324) (4339,3627) (2474,2204)
  };
  \addplot[
    only marks,
    mark=square*,
    mark size=0.5pt,
    color=myorange,
    fill=myorange,
  ] coordinates {
      (1296,1098) (75,56) (4252,3625) (1372,1183) (4543,3835) (3286,2968) (3132,2825) (3889,3399)
      (355,273) (4552,3833) (3565,3172) (2013,1812) (190,135) (3159,2827) (4335,3679) (1401,1202)
      (1743,1502) (3105,2774) (4159,3572) (259,160) (2265,2011) (4631,3880) (2178,1934) (2463,2184)
      (1635,1407) (3638,3199) (673,552) (4674,3900) (3788,3312) (2094,1885) (3768,3317) (4234,3607)
      (919,750) (1666,1444) (1434,1242) (3739,3294) (549,470) (3650,3226) (3428,3068) (2481,2175)
      (2283,2041) (3725,3279) (3094,2793) (2863,2549) (812,673) (2669,2375) (311,238) (4164,3560)
      (143,104) (2270,2029) (1781,1567) (3438,3082) (3311,2979) (3717,3277) (3310,3007) (357,290)
      (3884,3400) (3128,2788) (4580,3860) (1112,935) (4563,3844) (2499,2202) (210,189) (3455,3089)
      (3962,3456) (4291,3654) (1223,1043) (372,303) (686,563) (4593,3851) (3030,2699) (56,46)
      (602,494) (1671,1463) (1008,822) (4415,3730) (3906,3425) (376,305) (2751,2416) (266,193)
      (3366,3055) (1118,926) (2482,2187) (3517,3138) (1284,1102) (1329,1131) (3503,3134) (2690,2370)
      (542,445) (1425,1233) (1201,1018) (4048,3510) (406,335) (4287,3651) (381,310) (4492,3782)
      (3338,2996) (1604,1372) (2998,2677) (679,571) (684,563) (4543,3819) (1555,1345) (2392,2129)
      (1812,1558) (2373,2102) (2293,2043) (2368,2095) (4276,3641) (3830,3360) (1358,1180) (3790,3314)
      (4496,3779) (1349,1161) (358,290) (1550,1356) (4253,3626) (324,260)
  };
  \addplot[
    only marks,
    mark=square*,
    mark size=0.5pt,
    color=mypurple,
    fill=mypurple,
  ] coordinates {
      (4613,3882) (1597,1369) (591,513) (1080,910) (3427,3076) (2167,1918) (3259,2949) (1287,1121)
      (2140,1927) (2310,2075) (970,790) (4506,3792) (1652,1424) (3460,3093) (903,740) (2192,1957)
      (4664,3906) (2661,2371) (488,428) (2569,2290) (631,547) (2299,2065) (1732,1489) (13,5)
      (4671,3910) (618,531) (737,623) (2025,1828) (3256,2945) (1750,1529) (2888,2606) (2156,1915)
      (1656,1429) (1858,1646) (2281,2036) (1795,1568) (3360,3052) (823,681) (901,742) (1598,1371)
      (58,45) (1062,864) (4573,3835) (837,706) (46,37) (1661,1433) (142,125) (1988,1779)
      (3417,3072) (504,447) (4484,3771) (1583,1361) (1898,1674) (375,334) (3184,2881) (4654,3896)
      (4505,3796) (1001,814) (1805,1552) (624,528) (1969,1751) (437,388) (2107,1891) (1775,1539)
      (3219,2916) (3288,2982) (4612,3882) (1380,1200) (3564,3151) (1587,1360) (2749,2472) (2160,1915)
      (1977,1769) (1033,855) (71,60) (1390,1221) (1071,898) (1895,1671) (539,458) (687,597)
  };
  \node[anchor=center, font={\scriptsize\sffamily\bfseries}, rotate=90] at (axis cs:350.8,1958.5) {English doc};
  \node[anchor=south, inner sep=1.5pt, font={\scriptsize\sffamily\bfseries}] at (axis cs:2154.14,3839.14) {Japanese doc};
\end{axis}
\end{tikzpicture}
        \caption{\textbf{\MDPAlign{}} (ours)}
        \label{fig:mdp}
    \end{subfigure} &
    \begin{subfigure}[t]{0.47\columnwidth}
        \centering
\begin{tikzpicture}[trim axis left, trim axis right]
\begin{axis}[
  width=0.92\linewidth,
  scale only axis,
  axis equal image,
  axis background/.style={fill=black!2},
  xmin=-0.5,
  xmax=4683.5,
  ymin=-0.5,
  ymax=3917.5,
  y dir=reverse,
  xtick={0,1000,2000,3000,4000},
  xticklabels={0,1k,2k,3k,4k},
  ytick=\empty,
  yticklabels={},
  ymajorticks=false,
  tick label style={
    font={\scriptsize\sffamily},
    inner sep=0.5pt,
  },
  tick align=outside,
  xtick pos=bottom,
  ytick pos=left,
  enlargelimits=false,
  clip=true,
]
  \addplot[
    only marks,
    mark=square*,
    mark size=0.5pt,
    color=mylightorange,
    fill=mylightorange,
  ] coordinates {
      (841,726) (2509,2242) (1871,1656) (812,696) (3264,2942) (1221,1056) (269,257) (2524,2257)
      (2790,2534) (2388,2138) (1232,1064) (2840,2576) (2307,2060) (4317,3609) (495,442) (322,296)
      (294,280) (3058,2764) (1643,1414) (194,189) (3259,2948) (3619,3172) (3647,3177) (1374,1204)
      (1473,1287) (612,540) (4261,3555) (3134,2848) (3283,2963) (2677,2394) (3182,2883) (1926,1712)
      (759,656) (1022,831) (2693,2433) (3609,3161) (2690,2410) (545,486) (697,606) (2633,2354)
      (436,387) (674,570) (2631,2353) (1052,872) (207,210) (2812,2540) (2724,2457) (2659,2361)
      (216,203) (3095,2802) (2018,1835) (1625,1390) (523,455) (2283,2038) (457,422) (2323,2078)
      (221,199) (2237,1987) (2254,2002) (1187,976) (85,76) (1474,1288) (2328,2107) (3863,3291)
      (1003,812) (1352,1167) (32,20) (1339,1157) (913,709) (84,61) (1873,1653) (2358,2126)
      (159,173) (3932,3320) (1539,1297) (991,802) (1216,1039) (2846,2572) (761,653) (2991,2697)
      (2224,1975) (1655,1428) (880,734) (199,194) (1351,1168) (2321,2088) (2112,1893) (2955,2677)
      (1879,1649) (1897,1669) (3501,3108) (4372,3663) (2065,1857) (1193,1016) (3843,3279) (538,459)
      (687,598) (3351,3032) (3740,3225) (1922,1714) (1047,876) (2408,2168) (1020,834) (906,716)
      (1428,1234) (1990,1775) (3178,2886) (1562,1349) (1517,1313) (1854,1642) (2384,2141) (2674,2392)
      (338,303) (873,738) (3912,3310) (145,155) (2521,2266) (521,452) (471,433) (869,738)
      (2678,2394) (1913,1700) (3472,3099) (2658,2361) (992,802) (3833,3272) (2689,2413) (1902,1686)
      (4383,3673) (2290,2049) (564,492) (1581,1363) (2791,2534) (4326,3619) (2359,2126) (3326,3012)
      (402,362) (2407,2168) (2581,2321) (81,64) (2040,1813) (2770,2517) (1343,1153) (2389,2138)
      (2950,2666) (620,534) (1211,1023) (734,618) (1818,1560) (765,651) (3535,3132) (3940,3322)
      (1064,863) (2819,2538) (2549,2284) (2685,2408) (2719,2458) (1127,959) (1580,1363) (2853,2569)
      (2198,1935) (3274,2957) (990,803) (2288,2051) (2872,2616) (4302,3596) (4384,3673) (573,502)
      (3031,2745) (1304,1148) (1862,1663) (12,4) (2808,2548) (4115,3414) (2878,2614) (82,63)
      (1391,1220) (1901,1688) (783,665) (4282,3575) (1924,1707) (3444,3083) (3987,3347) (2693,2426)
      (2950,2667) (609,540) (3108,2816) (1537,1297) (3435,3080) (2593,2312) (1785,1579) (184,161)
      (821,692) (1760,1515) (290,284) (4,0) (1637,1420) (2547,2284) (3295,2988) (3748,3227)
      (2539,2271) (708,582) (2286,2047) (1842,1601) (1252,1101) (3865,3284) (572,496) (257,228)
      (1714,1460) (1920,1713) (3352,3032) (2176,1931) (828,680) (3320,3015) (1599,1371) (1086,903)
      (462,416) (2637,2352) (1829,1626) (1843,1600) (3256,2945) (894,748) (3673,3190) (3273,2957)
      (4260,3554) (1361,1180) (3316,3021) (696,608) (2617,2335) (4362,3653) (3430,3077) (2577,2306)
      (1796,1565) (1877,1649) (2780,2525) (3616,3170) (4279,3571) (942,767) (1057,886) (914,709)
      (820,692) (2444,2191) (2564,2294) (3833,3271) (2913,2646) (964,784) (3550,3135) (1901,1690)
      (2290,2048) (4320,3613) (2953,2674) (1396,1218) (3060,2774) (3116,2833) (1983,1783) (3078,2784)
      (4626,3879) (416,370) (2778,2526) (1920,1707) (990,804) (3099,2800) (389,338) (4167,3460)
      (1373,1204) (163,179) (2951,2667) (2527,2256) (85,77) (3306,3001)
  };
  \addplot[
    only marks,
    mark=square*,
    mark size=0.5pt,
    color=myorange,
    fill=myorange,
  ] coordinates {
      (353,318) (3140,2857) (4343,3648) (447,375) (3138,2849) (4631,3880) (3489,3105) (4525,3812)
      (3070,2788) (1830,1626) (34,27) (1677,1447) (224,191) (552,474) (2828,2565) (20,15)
      (1307,1134) (4372,3661) (2923,2648) (3194,2901) (1380,1202) (3331,3034) (4444,3737) (1107,921)
      (679,571) (3517,3138) (2619,2337) (1364,1185) (972,793) (511,449) (2040,1819) (2128,1918)
      (3285,2963) (1537,1354) (1818,1562) (800,706) (4616,3884) (1976,1757) (2937,2661) (4434,3722)
      (4678,3913) (2018,1822) (707,611) (2071,1873) (3247,2938) (4496,3779) (2438,2187) (1471,1285)
      (635,550) (1063,865) (950,783) (2046,1854) (1637,1410) (3774,3240) (1795,1574) (1981,1789)
      (4001,3303) (2291,2052) (4459,3744) (3366,3055) (1078,903) (3806,3278) (994,810) (2018,1839)
      (1026,843) (2689,2404) (3252,2942) (216,219) (1787,1571) (159,160) (1131,940) (2885,2613)
      (425,381) (4364,3657) (4461,3751) (1707,1461) (288,278) (3495,3107) (2911,2631) (3107,2815)
      (875,738) (1122,939) (2779,2530) (1395,1219) (2793,2534) (426,393) (3220,2918) (4360,3652)
      (2370,2137) (4618,3829) (1357,1179) (3712,3202) (1650,1422) (1910,1699) (2118,1906) (282,262)
      (2872,2605) (3183,2891)
  };
  \addplot[
    only marks,
    mark=square*,
    mark size=0.5pt,
    color=mypurple,
    fill=mypurple,
  ] coordinates {
      (3679,3195) (4450,3726) (4373,3666) (4044,3369) (2975,2715) (4184,3470) (3160,2868) (3384,3059)
      (1750,1529) (4612,3882) (3287,2973) (1469,1281) (2266,2012) (1780,1591) (229,225) (250,232)
      (281,269) (1070,895) (2332,2113) (17,12) (1977,1769) (4392,3686) (34,21) (930,753)
      (355,319) (2956,2677) (4501,3786) (1489,1333) (4254,3548) (2684,2401) (4101,3407) (1487,1332)
      (3150,2864) (1311,1142) (2269,2027) (4252,3546) (613,538) (4248,3542) (521,453) (2186,1949)
      (2204,1958) (3888,3301) (3162,2875) (2220,1970) (1772,1536) (138,137) (3735,3218) (251,233)
      (1280,1130) (2483,2222) (488,428) (1444,1251) (3778,3242) (2310,2075) (1255,1096) (1384,1209)
      (1730,1497) (4470,3753) (2070,1862) (2165,1917) (3675,3197) (2577,2308) (3309,3018) (2299,2065)
      (4664,3906) (1651,1423) (247,236) (2113,1893) (4067,3388) (427,383) (42,36) (3173,2893)
      (3429,3078) (3044,2760) (261,244) (2822,2567) (3419,3073) (3057,2765) (737,623) (491,434)
      (1654,1428) (603,527) (1001,814) (2832,2582) (1458,1264) (3059,2774) (4271,3565) (453,407)
      (2952,2672) (4107,3410) (2675,2392) (214,201) (1202,1021) (1728,1492) (1516,1317) (835,685)
      (1909,1698) (2341,2120) (2186,1950) (3021,2743) (660,562) (4096,3401) (1939,1744) (4267,3561)
      (3990,3345) (4658,3899) (2023,1827) (3076,2785) (3645,3183) (4241,3534) (3197,2904) (3884,3303)
      (4335,3625) (1817,1561) (1884,1652) (1595,1368) (78,66) (746,626) (4277,3569) (4668,3903)
      (1147,966) (2855,2586) (1732,1489) (1484,1301) (1941,1739) (4411,3705) (29,19) (3887,3300)
      (3670,3192) (2967,2679) (268,256) (765,649) (4360,3651) (742,624) (4141,3429) (2400,2161)
      (1583,1361) (4479,3765) (4623,3879) (4030,3368) (4413,3707) (1930,1722) (1488,1333) (363,326)
      (3699,3201) (2117,1905) (4472,3762)
  };
\end{axis}
\end{tikzpicture}
        \caption{\textbf{\CTFAlign{}} (ours)}
        \label{fig:fig1-ctf}
    \end{subfigure}
    \end{tabular}
    }

    \caption{Performing word alignment on a pair of English and Japanese Wikipedia articles shows the differences between alignment algorithms and the \textcolor{mypurple}{correct alignments}, \textcolor{mylightorange}{false negatives}, and \textcolor{myorange}{false positives} they predict.
    \SimAlign{}~\cite{Jalili-Sabet2020-rb} produces some stray false positives that are implausible given the structure of the documents.
    \MDPAlign{} focuses on the diagonal but misses the bend towards the end; \CTFAlign{} is closest to the gold alignment.
    The figure is based on \texttt{Qwen3-Embedding-4B} embeddings and subsamples 10\% of points for visualization. The two axes are the token positions in the document pair.}
    \label{fig:fig1}
\end{figure}

Word alignment is a foundational task in natural language processing with the aim to identify semantically corresponding words between parallel texts. The task has applications such as hallucination detection~\citep{Zhou2021-zs, Li2022-mk, Huang2024-mz}, label projection~\citep{Ehrmann2011-cd, Garcia-Ferrero2022-fo, Chen2023-yy, Fetahu2023-mz}, machine translation~\citep{Kong2019-qn, Wu2024-tc, Koshkin2024-nt, Wang2025-vi}, and cross-lingual representation learning~\citep{Miao2024-mf}. 

Most of these tasks have moved beyond the sentence level with large language models (LLMs) and embedding models being capable of processing entire documents in a single pass~\citep{Zhang2024-oj, Marone2025-ah, Boizard2025-ye, Zhang2025-kx}. In contrast, word alignment methods still largely operate on sentence pairs~\citep{Dou2021-dp, Wu2023-vw, Huang2024-mz, Miao2025-sk}, despite the fact that embedding-based approaches such as \SimAlign{}~\citep{Jalili-Sabet2020-rb} are technically applicable to full documents. 

However, when sentence-level word alignment algorithms are directly applied to documents, performance degrades as similarity matrices become larger and noisier (Figure~\ref{fig:fig1}b). 
In this work, we investigate how embedding-based word alignment can be extended from sentences to full documents without relying on sentence segmentation or additional supervision.

Our contributions are summarized as follows:
\begin{itemize}
    \item We introduce two lightweight, model-agnostic methods for document-level word alignment: \MDPAlign{} (Figure~\ref{fig:fig1}c) and \CTFAlign{} (Figure~\ref{fig:fig1}d).
    
    \item Our approaches reduce the degradation caused by scaling sentence-level word alignment to full documents, with \CTFAlign{} in particular matching sentence-level performance in several settings.
    
    \item We show that our document-level word alignment approaches transfer to downstream tasks, improving translation coverage evaluation and token-level recognition of semantic differences.
\end{itemize}

\section{Background \& Related Work}

\subsection{Word Alignment}
\label{subsec:wa}

Modern word alignment approaches typically leverage neural networks to solve this task with varying levels of supervision: \SimAlign{}~\citep{Jalili-Sabet2020-rb} showed that contextual embeddings from multilingual encoders can be leveraged to produce competitive word alignments without further training by applying similarity-based matching algorithms. \citet{Dou2021-dp} subsequently showed that embedding-based word alignment algorithms perform better if the underlying encoder is fine-tuned with parallel data. \citet{Arase2023-md} instead formulate embedding-based word alignment as an optimal transport problem, including partial and unbalanced variants that account for null alignments. WSPAlign~\citep{Wu2023-vw} proposed to include supervision signals from parallel, word-aligned as well as noisy data, and BinaryAlign~\citep{Latouche2024-np} framed word alignment as a binary sequence labeling task. Most recently, word alignment has been approached by fine-tuning LLMs on word alignment labeled data~\citep{Miao2025-sk}. These methods consistently focus on word alignment between sentences.

Word alignment between documents remains comparatively underexplored. In practice, for longer texts the task would have to be broken down into multiple steps by segmenting texts either into sentences or chunks and aligning them before applying sentence-level word alignment. The dependence on sentence segmentation and sentence alignment can be brittle, especially for languages or domains where reliable segmentation and alignment tools are unavailable. It may also discard document-level information and introduce cascading errors before word alignment is even performed. Even where such a pipeline remains usable despite imperfect intermediate steps, it introduces additional engineering and maintenance overhead: embedding-based pipelines require segmentation, sentence representation and alignment, followed by token representation and word alignment. In contrast, with long-context encoders, our approach obtains token representations for the full document in a single encoder pass and performs alignment directly on the similarity matrix.

\subsection{Positional Constraints and Priors in Alignment}
Alignment methods have long relied on positional structure and ordering assumptions. Based on the observation that translations tend to preserve the order of the content, alignment was conditioned on relative token positions for IBM Model 2~\citep{Brown1993-ld,dagan-etal-1993-robust}. FastAlign~\citep{Dyer2013-ml} later simplified this idea into an efficient diagonal positional prior tailored to sentence-level word alignment, and \SimAlign{}~\citep{Jalili-Sabet2020-rb} proposed a distortion correction that penalizes large deviations in relative position.
Related assumptions have also been used in sentence alignment.~\citet{Gale1993-zn} exploited the near-monotonic structure of parallel documents by constraining alignment paths to remain diagonal.

\MDPAlign{} extends this intuition from sentence and sentence-alignment settings to document-level word alignment by imposing a global diagonal prior.

\subsection{Coarse-to-Fine Methods}
\label{subsec:ctf-prev}

Coarse-to-fine strategies have been widely used across NLP to reduce large search spaces by progressively refining predictions~\citep{Petrov2008-do, Petrov2011-ij, Dong2018-ra}. 
Closest to our work is VecAlign~\citep{Thompson2019-es}, a sentence alignment method that recursively narrows the alignment search space within a document using sentence embeddings. However, VecAlign operates on sentence representations and searches for a monotonic, contiguous alignment path with dynamic programming. In contrast, \mbox{\CTFAlign{}} applies coarse-to-fine refinement directly to token-level similarity matrices and remains compatible with non-monotonic, non-contiguous alignment patterns at all resolutions.

\begin{figure}[t]
    \centering

    \begin{subfigure}{0.5\columnwidth}
        \centering
        \includegraphics[width=\linewidth]{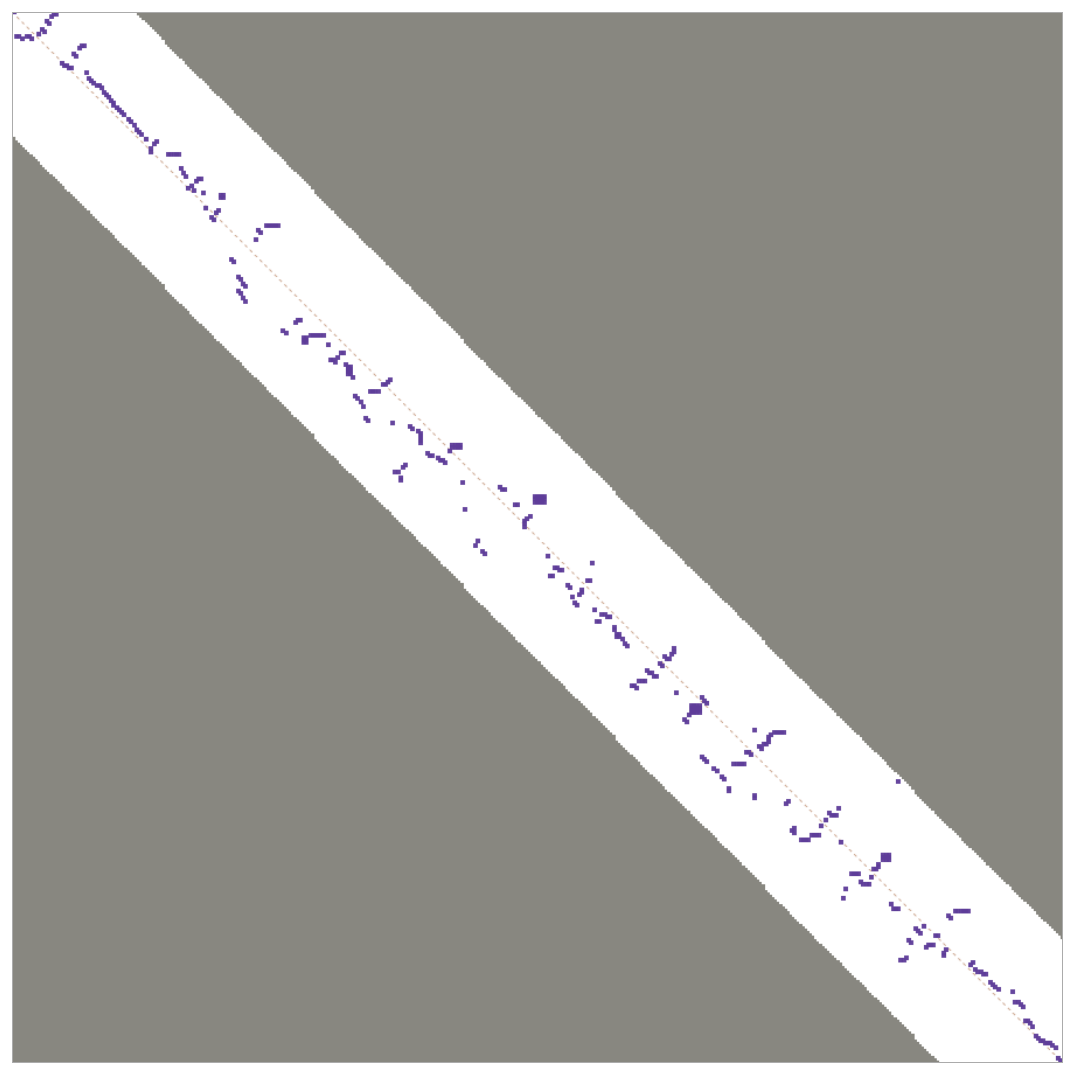}
        \caption{Strict \MDPAlign{} with bandwidth $k=50$.}
        \label{fig:mdp-strict}
    \end{subfigure}
    
    \vspace{2mm}
    
    \begin{subfigure}{0.5\columnwidth}
        \centering
        \includegraphics[width=\linewidth]{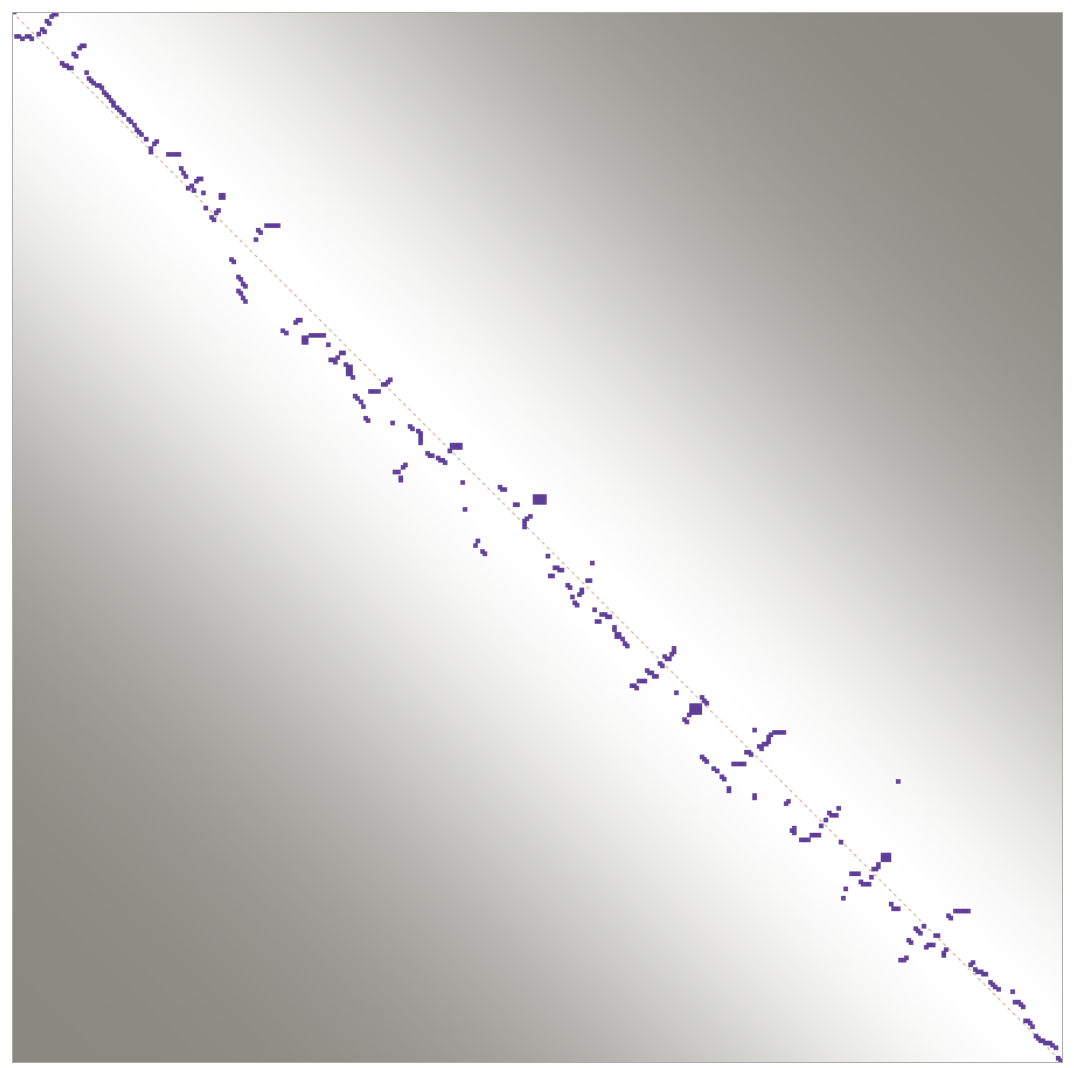}
        \caption{Fuzzy \MDPAlign{} masking with gradual decay away from the diagonal with $k=150$.}
        \label{fig:mdp-fuzzy}
    \end{subfigure}

    \caption{Illustration of the two \MDPAlign{} variants. Both assume similar relative positions of corresponding words across documents and constrain alignment search toward the main diagonal. \textcolor{mypurple}{Alignments} shown in search space with darkness of gray indicating suppression of off-diagonal similarities.}
    \label{fig:mdpalign}
\end{figure}

\begin{figure*}[t]
  \centering

  \begin{subfigure}{0.24\textwidth}
    \includegraphics[width=\linewidth]{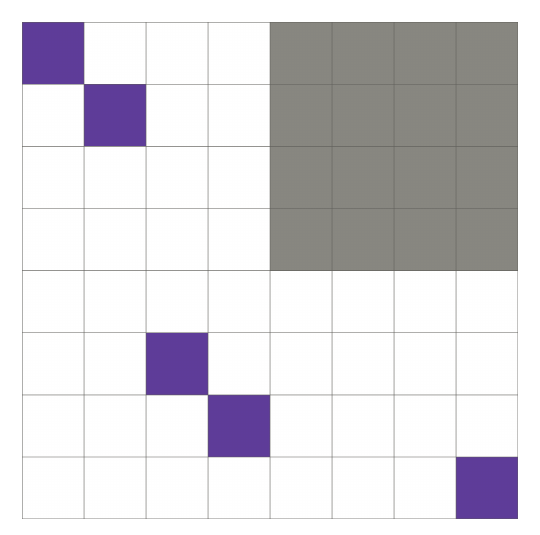}
    \caption{Detection of \textcolor{mypurple}{high-similarity blocks} in a coarse-grained grid using average-pooled cosine similarities. The upper-right region was masked at the previous resolution.}
  \end{subfigure}
  \hfill
  \begin{subfigure}{0.24\textwidth}
    \includegraphics[width=\linewidth]{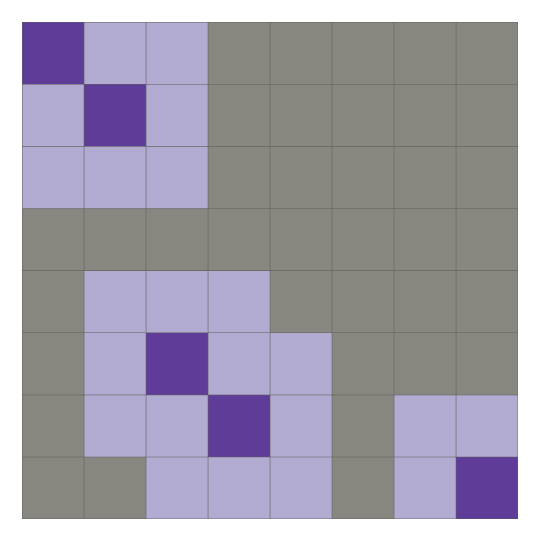}
    \caption{Expansion of detected blocks with a \textcolor{mylightpurple}{buffer} of $w=1$ to neighboring regions.\\[20pt]}
  \end{subfigure}
  \hfill
  \begin{subfigure}{0.24\textwidth}
    \includegraphics[width=\linewidth]{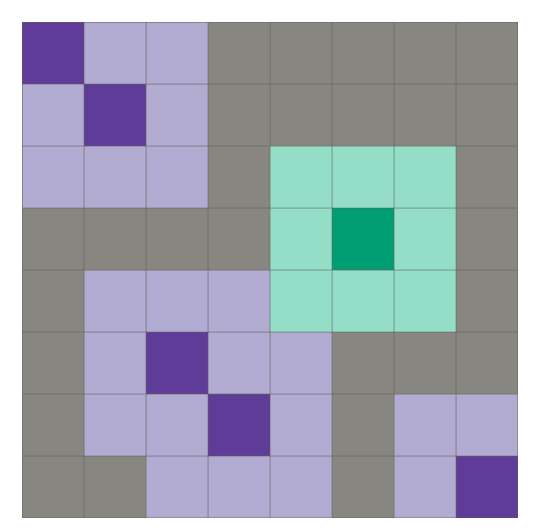}
    \caption{Recovery of \textcolor{mygreen}{blocks with fully empty rows and columns} using a \textcolor{mylightgreen}{buffer} of $w=1$.\\[20pt]}
  \end{subfigure}
  \hfill
  \begin{subfigure}{0.24\textwidth}
    \includegraphics[width=\linewidth]{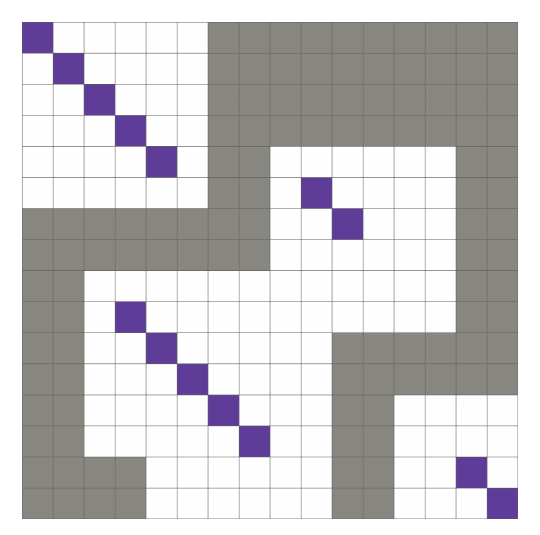}
    \caption{Refinement at higher resolution by halving block size and repeating the search in the remaining unmasked regions.\\[10pt]}
  \end{subfigure}

  \caption{Illustration of one iteration of the \CTFAlign{} algorithm. White regions denote searchable grid space, while \textcolor{mygray}{gray regions} are masked because they fall outside the aligned coarse blocks identified at the previous resolution. The process repeats from an initial $2 \times 2$ grid down to the token-level $m \times n$ grid.}
  \label{fig:ctf}
\end{figure*}

\section{Methods}
\label{sec:methods}

\subsection{Document-level Word Alignment}
\label{subsec:overview}

\paragraph{Task description}
Given a parallel document pair consisting of a source word sequence $\mathbf{x} = (x_1, \dots, x_m)$ and a target word sequence $\mathbf{y} = (y_1, \dots, y_n)$, the goal of document-level word alignment is to produce a set of alignment links $\mathcal{A} \subseteq \{1,\dots,m\} \times \{1,\dots,n\}$ connecting semantically corresponding words across the two documents.
Unlike sentence-level word alignment, documents may contain thousands of words, sentence boundaries may be unavailable, and lexical repetitions can induce many plausible but incorrect matches across the whole document. As a result, the alignment search space for documents becomes noisier and less locally constrained than for sentences.

\paragraph{Embedding-based alignment}
We build on \SimAlign{}~\citep{Jalili-Sabet2020-rb}, which extracts contextual token representations from a multilingual encoder and computes a cosine-similarity matrix $S \in \mathbb{R}^{m \times n}$, where each entry $S_{ij}$ represents the similarity between source token $x_i$ and target token $y_j$. 
Alignment links are then predicted based on $S$ using an alignment algorithm such as Argmax or Itermax~\citep{Jalili-Sabet2020-rb}. Possible but rare negative cosine similarities are not considered for alignments. Both identify high-similarity token pairs bidirectionally: Argmax does so in a single pass, leading to sparser alignments, while Itermax applies the procedure iteratively to recover additional links. The token-level alignments are then converted to word alignments by mapping each subword token back to its parent word and treating two words as aligned whenever at least one of their token pairs is linked.
While effective at the sentence level, if applied without constraint, these algorithms become susceptible to noise in document-length similarity matrices and lead to spurious alignments, as shown in Figure~\ref{fig:fig1}b.

\paragraph{Structural constraints for document-level alignment}
Our central hypothesis is that the primary challenge in document-level word alignment lies in the unconstrained search space of the similarity matrix. We therefore propose to constrain the alignment search space by applying structural constraints directly to the similarity matrix before applying an alignment algorithm.

We introduce two mechanisms for constraining the document-level word alignment search space. First, \textbf{\MDPAlign{}} (Subsection~\ref{subsec:mdp}; Figure~\ref{fig:mdpalign}) builds on the assumption that translations often preserve global structure, such that corresponding content occupies approximately the same relative position in both documents. It encodes this prior through diagonal-band masking, biasing alignments toward the main diagonal of the similarity matrix.

While this positional correspondence oftentimes holds for direct translations, realistically, documents that convey the same content may also exhibit reordering, insertions, and deletions. We therefore introduce as a second approach \mbox{\textbf{\CTFAlign{}}} (Subsection~\ref{subsec:ctf}; Figure~\ref{fig:ctf}), which performs adaptive coarse-to-fine refinement, recursively identifying likely alignment regions and restricting subsequent alignment to those regions, allowing the alignment structure to adapt throughout the document. 

Both methods operate as lightweight, model-agnostic processing steps of the similarity matrix and require no further training of the underlying encoder.

\subsection{Main Diagonal Prior (\MDPAlign{})}
\label{subsec:mdp}
Let $S \in \mathbb{R}^{m \times n}$ be the token-level similarity matrix for a source of $m$ tokens and a target of $n$ tokens. As shown in Figure~\ref{fig:mdpalign}, we map source token $i$ to its relative position $i/m$ and target token $j$ to $j/n$, and define their relative distance from the main diagonal as

\begin{equation}
    d_{ij} = \left|\frac{i}{m} - \frac{j}{n}\right|.
\end{equation}

\noindent{}We then apply one of two masking schemes, controlled by a bandwidth hyperparameter $k$. We anchor $k$ to the longer document as an absolute token count rather than a fraction of document length, on the assumption that reordering in well-translated text is largely local (clause- or sentence-level) and does not grow with document length.

\paragraph{Strict} We zero out all entries outside a diagonal band:

\begin{equation}
\tilde{S}_{ij} =
\begin{cases}
S_{ij}, & \text{if } d_{ij} \leq \frac{k}{\max(m,n)}, \\
0, & \text{otherwise}.
\end{cases}
\end{equation}

\paragraph{Fuzzy} Since zeroing out all entries beyond a certain band is oftentimes too harsh, we alternatively apply a Gaussian decay, where $k$ determines the standard deviation, preserving every cell but suppressing off-diagonal scores exponentially:

\begin{align}
    \tilde{S}_{ij}
    &=
    S_{ij}
    \cdot
    \exp\!\left(
        -\frac{d_{ij}^{2}}{2\sigma^{2}}
    \right), \\
    \sigma
    &=
    \frac{k}{\max(m,n)}.
\end{align}

\noindent{}To the modified matrix $\tilde{S}$ we then apply the \SimAlign{}~\citep{Jalili-Sabet2020-rb} alignment algorithm.\footnote{\citep{Jalili-Sabet2020-rb} also introduce a variant of \SimAlign{} with a parabolic discount for distortion control. However, they report mixed results on the sentence-level, and we deem their parabolic discount, flatter than our Gaussian one, unsuitable for long documents.}


\subsection{Coarse-to-Fine Alignment (\CTFAlign{})}
\label{subsec:ctf}

As visualized in Figure~\ref{fig:ctf}, we divide the similarity matrix $S$ into a $2 \times 2$ coarse grid by average-pooling cosine similarities, where negative similarities count as zero, and run \SimAlign{}~\citep{Jalili-Sabet2020-rb} on this coarse matrix. Each aligned block pair identifies a region of $S$ likely to contain true alignments. We zero out all regions of $S$ not covered by any aligned block. Formally, at each intermediate level $t$:

\begin{equation}
S^{(t+1)}_{ij} =
\begin{cases}
S^{(t)}_{ij}, & \text{if } (i,j) \in \mathcal{M}^{(t)}, \\
0, & \text{otherwise}.
\end{cases}
\end{equation}

\noindent{}where $\mathcal{M}^{(t)}$ is the set of token pairs whose containing coarse block was retained at level $t$. We then halve the block size and repeat on $S^{(t+1)}$, roughly doubling the resolution at each step, until we reach token resolution, at which point the alignment produced on the final $m \times n$ grid is returned directly. If document lengths are skewed, token resolution is reached for one side earlier than for the other. In these cases, the shorter side is held at token resolution, while the other continues to be refined, so the process terminates only once both sides have reached token resolution.

Two mechanisms improve robustness against coarse-level errors. First, a width hyperparameter $w \geq 0$ controls how many neighboring blocks are retained around each aligned block at every resolution. Second, we apply a recovery step at each level: any block at the intersection of a source row and a target column, of which neither has any other retained block, is recovered. A recovered block also retains a buffer of width $w$ additional blocks around it. A reference implementation of the \CTFAlign{} algorithm is provided in Appendix~\ref{app:pseudo}.

\section{Experimental Setup}
\label{sec:experimental-setup}

\paragraph{Data} Since there are currently no datasets that include word alignment annotation directly for documents, we reconstruct documents from human-annotated sentence-level word alignment datasets that include information about document membership and sentence order. Hence, our evaluation includes English--French (en--fr;~\citeauthor{Mihalcea2003-hd},~\citeyear{Mihalcea2003-hd}), English--Romanian (en--ro;~\citeauthor{Mihalcea2003-hd},~\citeyear{Mihalcea2003-hd}), English--Japanese (en--ja;~\citeauthor{neubig11kftt},~\citeyear{neubig11kftt}), English--Chinese (en--zh;~\citeauthor{Liu2015-zz},~\citeyear{Liu2015-zz}), English--Czech (en--cz;~\citeauthor{Marecek2011-nm},~\citeyear{Marecek2011-nm}), and Latin--Greek (la--gr;~\citeauthor{Yousef2022-jg},~\citeyear{Yousef2022-jg})---covering six language pairs with varying degrees of typological distance and resourcedness. We use existing development/test splits where available. For la--gr, we split 25\% into a development set and leave the remaining 75\% as a test set, while en--cz is only considered as a test set. Statistics on datasets and splits are presented in Appendix~\ref{app:dataset}. 

\paragraph{Models} We generate contextualized token embeddings for entire document pairs using two recent multilingual long-context models that encode documents in a single forward pass: \texttt{Qwen3-Embedding-4B} (40,960 tokens;~\citeauthor{Zhang2025-kx},~\citeyear{Zhang2025-kx}), a decoder-based embedding model contrastively trained on large-scale text pairs, including bitext-mining data, and \texttt{mmBERT-base} (307M parameters, 8,192 tokens;~\citeauthor{Marone2025-ah},~\citeyear{Marone2025-ah}), a general-purpose multilingual encoder pretrained with masked language modeling without explicit parallel-data or pairwise embedding supervision. Additionally, we include \texttt{LaBSE} (500M parameters;~\citeauthor{Feng2022-oc},~\citeyear{Feng2022-oc}), which is explicitly trained for cross-lingual similarity but has a more limited context window (512 tokens). In this case, we embed sentences individually and concatenate their representations. Model choice, layer selection, and hyperparameters are determined on the development set. Development results are reported in Appendix~\ref{app:dev}. For language pairs without a development split (i.e., en--cz), we choose the layer that performed best most frequently across the remaining language pairs at the same granularity. We then compute token-level cosine similarity matrices from the resulting token embeddings.

\paragraph{Baselines} We use \SimAlign{}~\citep{Jalili-Sabet2020-rb} with Argmax as our baseline at both sentence- and document-level.\footnote{For our experiments with the development set, we also considered Itermax as a baseline. See Appendix~\ref{app:dev} for details.} 

The unconstrained document-level setting serves as our primary baseline for measuring the effect of document-scale alignment, while the sentence-level setting provides an approximate upper bound.

As an additional baseline, we use an LLM to autoregressively generate the word alignment labels given a simple 1-shot prompt without fine-tuning. Our prompt is based on the ``full-mode'' prompt defined by ~\citet{Miao2025-sk}, which generates alignment labels in one pass. We disregard their ``marker mode'' prompt, which would rely on as many passes as both documents contain words, deeming it unfeasible for long documents. We use \texttt{gpt-5.4-mini-2026-03-17} to generate the JSON output in OpenAI's structured output mode with a temperature of 0, while all other parameters are kept at default (see Appendix~\ref{app:llm-baseline} for the prompt). 

\begin{figure*}[t]
  \centering
  \includegraphics[width=\textwidth]{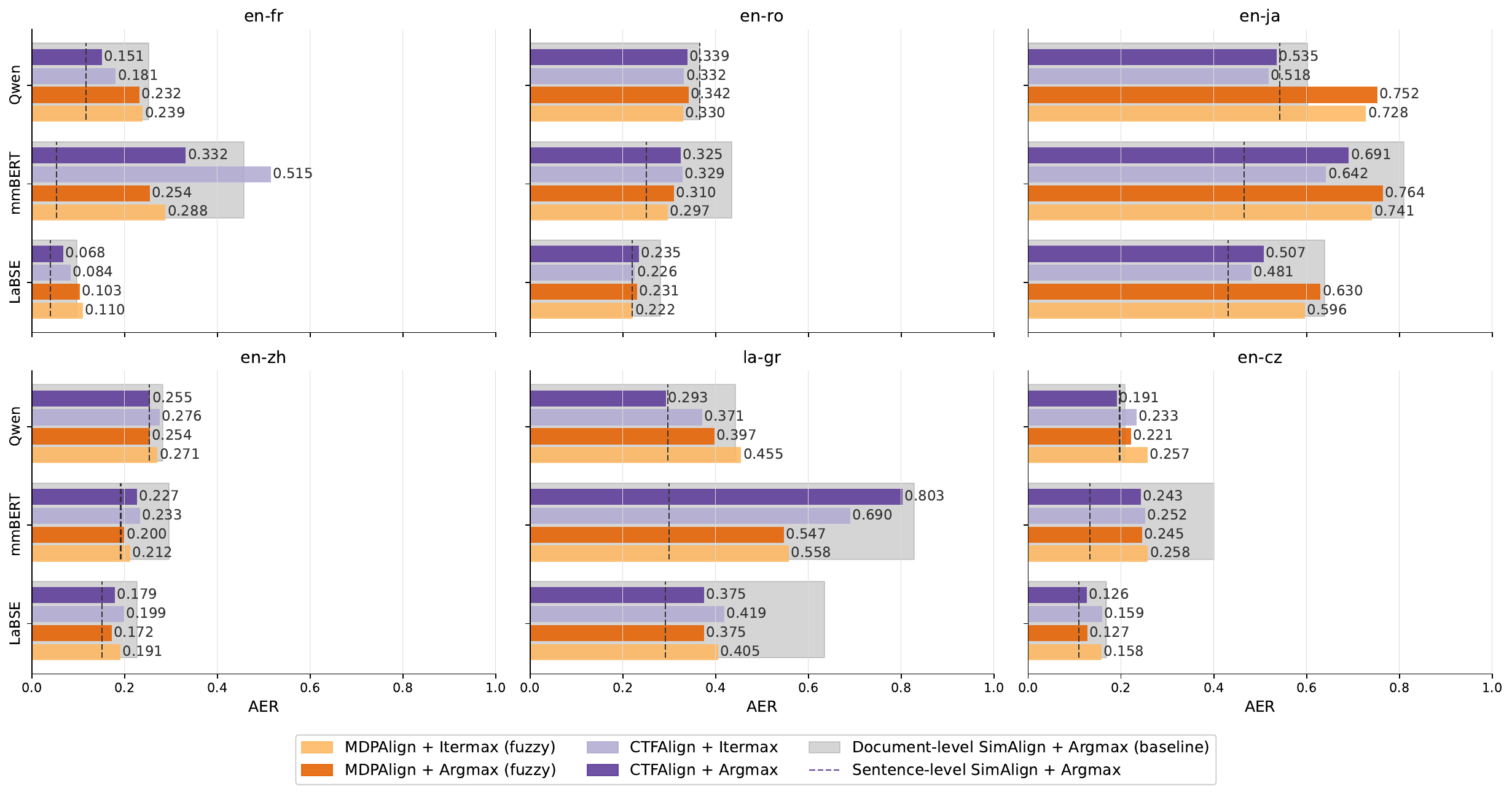}
  \caption{AER (lower is better) on the test set with \texttt{Qwen3-Embedding-4B},  \texttt{mmBERT-base}, and \texttt{LaBSE}. Different constrained word alignment approaches are denoted by different colors, the document-level baseline is shown in gray, and the dashed lines indicate sentence-level \SimAlign{} performance of the given model.}
  \label{fig:test-results}
\end{figure*}

\paragraph{Implementation Details} 
We evaluate both variants of \MDPAlign{} (\textit{strict} and \textit{fuzzy}) as well as \CTFAlign{} in combination with both Argmax and Itermax on the development set. We perform hyperparameter searches over the parameters $k$ and $w$ (Appendix~\ref{app:hyp-selection}). Based on best development set performance (Appendix~\ref{app:dev-res}), we select \MDPAlign{} \textit{fuzzy} with $k=150$ and \CTFAlign{} with $w=8$ for evaluation on the test set.

\paragraph{Evaluation} We follow word alignment evaluation from prior work~\citep{Jalili-Sabet2020-rb, Dou2021-dp, Miao2025-sk} and report alignment error rate (AER) on the test set in Figure~\ref{fig:test-results} for \texttt{Qwen3-Embedding-4B},  \texttt{mmBERT-base}, and \texttt{LaBSE}. In Appendix~\ref{app:full-results}, we include F1, Precision, Recall, and corresponding macro averages. Find formalizations of the metrics in Appendix~\ref{app:eval-formulas}.

\begin{table}[t]
\resizebox{\columnwidth}{!}{
\begin{tabular}{@{}l r r r r r r@{}}
\hline
\textbf{Granularity} & \textbf{en-fr} & \textbf{en-ro} & \textbf{en-ja} & \textbf{en-zh} & \textbf{la-gr} & \textbf{en-cz} \\
\hline
Sentences & 0.185 & 0.437 & 0.628 & 0.399 & 0.179 & 0.344 \\
Documents & 0.995 & 0.968 & 0.986 & 0.935 & 0.996 & 0.982 \\
\hline
\end{tabular}}
\caption{GPT-5.4-mini word alignment AER on the test set.}
\label{tab:gpt-results}
\end{table}

\section{Results \& Discussion}
\label{sec:results}

\paragraph{Unconstrained document-level search causes alignment collapse} Figure~\ref{fig:test-results} shows that across all models and language pairs, directly applying \SimAlign{} to full documents (gray areas) leads to a clear degradation in alignment quality compared to sentence-level performance (dashed lines). This effect becomes especially pronounced for long documents (en--fr, en--ja, la--gr) with \texttt{mmBERT}, where AER values can exceed 0.8 (en--ja, la--gr) at the document level. Manual inspection suggests that many of the resulting spurious alignments involve named entities that are repeated across the document as well as conjunctions, prepositions, and punctuation marks.

\paragraph{LLM prompting does not scale to document-level word alignment} Our experiments with \texttt{GPT-5.4-mini} reveal that autoregressively generating word alignment labels does not scale to documents (Table~\ref{tab:gpt-results}): while sentence-level performance is competitive on some language pairs (e.g., la--gr, en--fr), AER collapses to near-1.0 across all six pairs once the model is given full documents. These observations suggest that, at least without fine-tuning on document-level alignment data, similarity-matrix-based approaches remain better suited for this task than 1-shot prompting LLMs. This is consistent with~\citet{Miao2025-sk}, who report that LLM-based word alignment becomes competitive only after supervised fine-tuning.

\paragraph{Structural constraints recover document-level word alignment quality}
Both of our structural constraints recover most of the performance loss introduced by applying \SimAlign{} directly to full documents, in several settings closing the sentence- vs. document-level gap entirely. At the individual language-pair level, the gap narrows to within a few thousandths AER: en--ro with \texttt{LaBSE}+\MDPAlign{}+Itermax (0.222 document- vs. 0.220 sentence-level), and en--zh with \texttt{mmBERT}+\MDPAlign{}+Argmax (0.200 vs. 0.192). For \texttt{Qwen}, we even see document-level word alignment surpass sentence-level word alignment for several pairs: en--ro across all configurations, en--ja with \CTFAlign{}, la--gr and en--cz with \CTFAlign{}+Argmax.

\paragraph{\MDPAlign{} is effective when alignments remain close to the diagonal}
\MDPAlign{} improves over the unconstrained document-level baseline on document pairs whose alignment paths stay close to the global diagonal. However, the fixed diagonal prior cannot accommodate document pairs whose alignments do not follow this pattern. This is most visible on en--ja, where \MDPAlign{} falls under the unconstrained baseline (\texttt{Qwen}+\MDPAlign{} 0.752/0.728 vs. \SimAlign{} 0.602). Visualizations of the en--ja gold labels uncover curved alignment paths that deviate from the main diagonal and are therefore systematically penalized by \MDPAlign{} (Figure~\ref{fig:fig1}c). 

\paragraph{\CTFAlign{} remains robust across varying alignment structures}
\CTFAlign{}'s performance is comparable to \MDPAlign{} on pairs where the diagonal prior is well-suited (en--ro, en--zh), but \CTFAlign{}'s recursive refinement of the search space also adapts to curved or shifted alignment regions (Figure~\ref{fig:fig1}d). This leads to the strongest results where flexibility matters: on en--ja with \texttt{LaBSE} \CTFAlign{}+Itermax reaches 0.481 AER (closing more than two thirds of the gap to the sentence-level baseline), on la--gr with \texttt{Qwen} \CTFAlign{}+Argmax meets the sentence-level upper bound (0.293 vs. 0.292), and on en--fr with \texttt{LaBSE} it achieves the strongest document-level result overall (0.068), despite en--fr being the longest document pair in the dataset.

\emph{Averaged across all six language pairs, \mbox{\CTFAlign{}+Argmax} is the most robust configuration. It brings down average AER from 0.341 to 0.248 with LaBSE, within 0.041 of the sentence-level upper bound (0.207), and closes the gap between sentence- and document-level word alignment with \texttt{Qwen} (0.294 vs. 0.295).}

\paragraph{Structural constraints interact with encoder characteristics}
While \CTFAlign{} achieves the best results with \texttt{LaBSE} and \texttt{Qwen}, with \texttt{mmBERT} it is outperformed on average by \MDPAlign{} (0.437 vs.\ 0.387 AER for the Argmax variants). Unlike the other two encoders, \texttt{mmBERT} is a general-purpose pretrained encoder that has not been explicitly optimized with a pairwise embedding objective or parallel data. \CTFAlign{} still improves over unconstrained \texttt{mmBERT}, but less than \MDPAlign{}. This suggests that the two structural constraints interact differently with the representations produced by the underlying encoder. Since \CTFAlign{} recursively refines alignment regions, errors introduced at coarse resolutions can propagate to later stages, particularly with Itermax (see Appendix~\ref{app:fail} for visualizations). \MDPAlign{}'s fixed diagonal band sidesteps this failure mode, making it a safer choice when the encoder provides a weaker cross-lingual similarity signal---provided the alignment structure remains close to the main diagonal.

\paragraph{Pairwise cross-lingual supervision is associated with alignment quality}
Across nearly all settings, \texttt{LaBSE} remains the strongest embedding model despite being smaller and older than recent multilingual encoders. One possible explanation is exposure to pairwise cross-lingual supervision: \texttt{LaBSE}, which was explicitly trained on large-scale translation pairs, performs best, while \texttt{Qwen} undergoes contrastive embedding training on large-scale multi-task text pairs, including bitext-mining data. In contrast, \texttt{mmBERT} is pretrained as a general-purpose masked language model without parallel data or an explicit pairwise embedding objective. This observation aligns with prior work showing that \texttt{LaBSE} produces particularly well-structured multilingual representations for cross-lingual similarity tasks~\citep{Wang2022-xt, Dale2023-ts, Wastl2026-ku}. While the models differ along several dimensions beyond their training objectives and data, these results suggest that pairwise cross-lingual supervision may help produce alignment-friendly multilingual representations. This further motivates adapting recent multilingual encoders with parallel-data objectives~\citep{Dou2021-dp} specifically for word alignment.\footnote{We rule out the usage of gold sentence segmentation as the primary reason for \texttt{LaBSE}'s strong performance, since additional experiments with sentence embedding concatenation improve results with \texttt{mmBERT-base} only slightly and still leave them well below \texttt{LaBSE}'s (Table~\ref{tab:alignment-results-dev}).}

\section{Downstream Task Evaluation}

Beyond intrinsic alignment evaluation, we test whether our approaches transfer to two downstream alignment-based tasks: translation coverage evaluation (Subsection~\ref{subsec:coverage}) and recognition of semantic differences (RSD; Subsection~\ref{subsec:rsd}). Translation coverage evaluation uses token-level correspondences to identify omissions and additions (hallucinations) in machine translation output~\citep{Zhou2021-zs, Li2022-mk, Dale2023-zr, Huang2024-mz}, while RSD aims to measure semantic difference between tokens of comparable documents~\citep{Vamvas2023-zi, Wastl2026-ku, Lozano2026-bp}.

For both tasks we use DiffAlign~\citep{Vamvas2023-zi} as our baseline. DiffAlign assigns each token a dissimilarity score of $1 - \max_j S_{ij}$ from the subword cosine similarity matrix $S$. We adapt the structural priors from Section~\ref{sec:methods} to this scoring setup and use the best-performing long-context encoder from our alignment experiments, \texttt{Qwen3-Embedding-4B} (layer 20). Word-level scores are obtained by averaging scores across constituent subword tokens.

\subsection{Translation Coverage Evaluation}
\label{subsec:coverage}

\paragraph{Data \& Evaluation} For our experiments on translation coverage evaluation, we use document-level translation outputs from the WMT24 and annotations from WMT24 Metrics Shared Task~\citep{Kocmi2024-fw, Freitag2024-vz}.

The data consists of multi-sentence source texts translated by participating systems. We focus on English--German (en--de) and Japanese--Chinese (ja--zh)\footnote{We exclude English--Spanish because its annotations differ from the remaining language pairs.}, which have been annotated by professional translators using the Multidimensional Quality Metrics framework (MQM;~\citeauthor{Freitag2021-sk},~\citeyear{Freitag2021-sk}). This framework includes multiple error categories including Accuracy/Omission and Accuracy/Addition, which we consider, while discarding the other categories. We reconstruct the segments into documents and convert the span annotations into binary token-level labels (1 = addition/omission, 0 = everything else). Dataset statistics are shown in Appendix~\ref{app:wmt}. For evaluation, we follow~\citet{Dale2023-zr, Dale2023-ts} and use ROC AUC.

\paragraph{Results}
Table~\ref{tab:auc-results-token} shows that use of DiffAlign with structural constraints improves over the unconstrained baseline at segment- and document-level. Consistent with the word alignment results, DiffAlign\textsubscript{CTF+Argmax} emerges as the strongest configuration at the document level on both language pairs and both error types (ja--zh: +0.055 (omissions), +0.008 (additions); en--de: +0.036 (omissions), +0.02 (additions)). The segment- to document-level degradation is smaller than in the alignment experiments, likely because the smaller unit here is a segment pair, which spans multiple sentences rather than an individual sentence. \MDPAlign{} remains effective for the shorter and likely more rigidly structured segments, while \CTFAlign{} gains the advantage as document-level structural complexity increases.

\begin{table}
\resizebox{\columnwidth}{!}{
  \begin{tabular}{@{}l rr rr@{}}
    \hline
    & \multicolumn{2}{c}{\textbf{ja-zh}} & \multicolumn{2}{c}{\textbf{en-de}} \\
    \cmidrule(lr){2-3}\cmidrule(lr){4-5}
    \textbf{Method} & \textbf{Om.} & \textbf{Add.} & \textbf{Om.} & \textbf{Add.} \\
    \hline
    \multicolumn{5}{l}{\textsc{Segment-level}} \\
    \hspace{1em}\textit{DiffAlign} & 0.758 & 0.769 & 0.856 & 0.877 \\
    \hspace{1em}\textit{DiffAlign\textsubscript{MDP+Argmax}} & 0.789 & \textbf{0.783} & 0.867 & \textbf{0.885} \\
    \hspace{1em}\textit{DiffAlign\textsubscript{CTF+Argmax}} & \textbf{0.801} & 0.779 & \textbf{0.869} & 0.882 \\
    \hspace{1em}\textit{DiffAlign\textsubscript{CTF+Itermax}} & 0.781 & 0.776 & 0.862 & 0.880 \\[3pt]
    \multicolumn{5}{l}{\textsc{Document-level}} \\
    \hspace{1em}\textit{DiffAlign} & 0.733 & 0.763 & 0.831 & 0.843 \\
    \hspace{1em}\textit{DiffAlign\textsubscript{MDP+Argmax}} & 0.762 & 0.762 & 0.857 & 0.854 \\
    \hspace{1em}\textit{DiffAlign\textsubscript{CTF+Argmax}} & \textbf{0.788} & \textbf{0.771} & \textbf{0.867} & \textbf{0.863} \\
    \hspace{1em}\textit{DiffAlign\textsubscript{CTF+Itermax}} & 0.770 & 0.769 & 0.853 & 0.854 \\
    \hline
  \end{tabular}
}
\caption{Token-level ROC AUC for translation coverage evaluation with \texttt{Qwen3-Embedding-4B}.}
\label{tab:auc-results-token}
\end{table}

\begin{table}[t]
\centering
\resizebox{0.85\columnwidth}{!}{
  \begin{tabular}{@{}lrrr@{}}
    \hline
    \textbf{Method} & \textbf{en--de} & \textbf{en--fr} & \textbf{en--it} \\
    \hline
     \textit{DiffAlign}& 0.286 & 0.156 & 0.318 \\
     \textit{DiffAlign\textsubscript{MDP+Argmax}}   & 0.312 & 0.158 & 0.312 \\
     \textit{DiffAlign\textsubscript{CTF+Argmax}} & \textbf{0.330} & \textbf{0.189} & \textbf{0.353} \\
     \textit{DiffAlign\textsubscript{CTF+Itermax}} & 0.319 & 0.172 & 0.344 \\
    \hline
  \end{tabular}
}
\caption{Test set Spearman $\rho$ for SwissGov-RSD with \texttt{Qwen3-Embedding-4B}.}
\label{tab:diffalign-spearman-test}
\end{table}

\subsection{Recognition of Semantic Differences}
\label{subsec:rsd}

\paragraph{Data \& Evaluation} We evaluate on SwissGov-RSD~\citep{Wastl2026-ku}, a human-annotated benchmark for token-level recognition of semantic differences between related cross-lingual documents drawn from Swiss government sources. 
The benchmark covers three language pairs English--German (en--de), English--French (en--fr), English--Italian (en--it). We follow their evaluation protocol and report Spearman correlations between predicted and gold token-level scores.

\paragraph{Results} Table~\ref{tab:diffalign-spearman-test} shows that the same pattern holds for recognition of semantic differences: search-space restriction improves over the \mbox{DiffAlign} baseline across all three language pairs (+0.044 for en--de, +0.033 for en--fr, +0.035 for en--it), with DiffAlign\textsubscript{CTF+Argmax} yielding the highest correlations with gold labels and establishing new state-of-the-art results on SwissGov-RSD.
Notably, \MDPAlign{} offers little improvement across languages on this task, even underperforming the baseline on en--it. This suggests that \CTFAlign{}'s recursive search-space restriction is particularly well-suited to this setting of imperfectly parallel documents, containing subtle semantic shifts as well as entirely omitted or added paragraphs, with gold annotations obtained directly at the document level.
\\

\noindent\emph{Across both downstream tasks, \mbox{\CTFAlign{}+Argmax} remains the best-performing configuration, suggesting that its gains transfer well to realistic alignment-based downstream applications.}

\section{Conclusion}
\label{sec:conclusion}

We introduce two lightweight, training-free approaches for document-level word alignment: \mbox{\MDPAlign{}}, which imposes a global positional prior, and \CTFAlign{}, which recursively restricts alignment to semantically plausible regions. Across six language pairs, both methods substantially reduce the degradation observed when moving from sentences to full documents, with \mbox{\CTFAlign{}} leading to the most robust performance across diverse document structures and language pairs.

These improvements transfer beyond intrinsic alignment evaluation. We show that \CTFAlign{} improves the downstream tasks of translation coverage evaluation and semantic difference recognition, demonstrating that document-level word alignment can benefit cross-lingual document comparison tasks without sentence segmentation, additional supervision, or model retraining.

\section*{Limitations}
\paragraph{Sentence-level nature of the word alignment dataset} Word alignment gold data was tailored to sentences and had to be reconstructed to documents. This has the following consequences:
\begin{itemize}
    \item Document structures remain largely monotonic, they may not contain reordering, insertions, or deletions.
    \item Sentence-level annotations may fail to capture document-level phenomena such as long-range reordering, cross-sentence dependencies, discourse-driven translation shifts, or omissions and additions spanning multiple sentences.
    \item Concatenation of documents led to a small number of document pairs to evaluate on, e.g., the en--fr and la--gr test sets each consist of only a single document pair.
\end{itemize}

These limitations further motivate the inclusion of downstream document-level evaluations. In particular, while the WMT-based data was reconstructed from segment-level annotations, annotators were provided with broader contextual information during annotation. Moreover, SwissGov-RSD was annotated directly at the document level and therefore captures document-level semantic phenomena more naturally.

\paragraph{Dependency on encoder representations}
Our approaches operate directly on token similarity matrices and therefore depend on the representations produced by the underlying multilingual encoder. When cross-lingual token similarities are insufficiently distinctive, structural constraints cannot fully compensate, and iterative refinement strategies such as \CTFAlign{} may propagate errors introduced at earlier stages. At the same time, this limitation suggests a promising direction for future work: long-context multilingual encoders could be further optimized to produce representations better suited for fine-grained cross-lingual alignment.

\section*{Ethical Considerations}
No specific ethical risks have been identified for this work.
The artifacts used in this work, except for the proprietary LLM, have open licenses and due to their origin, the datasets are unlikely to contain personally identifiable information or offensive content.

\section*{Acknowledgments}
This work was funded by the Swiss National Science Foundation (project InvestigaDiff; no.~10000503). Additionally, we would like to thank Marius Huber for helpful feedback.

\bibliography{custom,paperpile}

\clearpage
\appendix

\clearpage
\section{\CTFAlign{} Reference Implementation}
\label{app:pseudo}

\begin{minipage}{\textwidth}
\begin{lstlisting}[style=pythonstyle]
def ctfalign(S, align, width):
    """
    S: m x n similarity matrix (source tokens x target tokens)
    align: base word alignment function, e.g. SimAlign
           argmax/itermax, returns a set of (row, col)
           index pairs
    width: buffer width w (in coarse blocks) kept around
           each aligned block
    """
    m, n = S.shape
    block_h = max(m // 2, 1)   # start from a 2x2 grid
    block_w = max(n // 2, 1)

    while block_h >= 1 and block_w >= 1:
        # 1. Coarsen: average-pool S into blocks
        coarse = average_pool(S, block_h, block_w)  # R x C
        R, C = coarse.shape

        # 2. Align on the coarse (block-level) matrix
        coarse_alignment = align(coarse)  # set of (i, j)

        # 3. Stop once blocks are single tokens
        if block_h == 1 and block_w == 1:
            return coarse_alignment

        # 4. Mark retained blocks: aligned block +- width
        mask = zeros(R, C)
        for (i, j) in coarse_alignment:
            r0, r1 = max(i - width, 0), min(i + width + 1, R)
            c0, c1 = max(j - width, 0), min(j + width + 1, C)
            mask[r0:r1, c0:c1] = 1

        # 5. Recovery: empty rows/cols get intersecting
        #    cells recovered, also with +- width buffer
        empty_rows = [i for i in range(R)
                      if mask[i, :].sum() == 0]
        empty_cols = [j for j in range(C)
                      if mask[:, j].sum() == 0]
        for i in empty_rows:
            for j in empty_cols:
                r0, r1 = max(i - width, 0), min(i + width + 1, R)
                c0, c1 = max(j - width, 0), min(j + width + 1, C)
                mask[r0:r1, c0:c1] = 1

        # 6. Zero out S wherever the mask is 0
        S = S * upsample(mask, block_h, block_w)

        # 7. Refine: halve block size, clamp at 1
        block_h = max(block_h // 2, 1)
        block_w = max(block_w // 2, 1)
\end{lstlisting}

\captionof{figure}{Simplified Python code for \CTFAlign{}. Tensor-level details
(average pooling and mask upsampling implemented via PyTorch's
\texttt{avg\_pool2d}/\texttt{interpolate}) are abstracted for readability.}
\label{fig:ctfalign-code}
\end{minipage}

\clearpage

\section{Evaluation Details}

\subsection{Dataset Statistics}
\label{app:dataset}

Table~\ref{tab:alignment-data} summarizes the statistics of the reconstructed document-level word alignment datasets used in our experiments. The datasets vary considerably in document length, number of sentences, and split sizes, covering both short texts (en--zh) and very long documents (en--fr, la--gr). 

\begin{table}[ht]
  \centering\small
  \setlength{\tabcolsep}{3pt}
  \begin{tabular}{l rr rr rr}
    \toprule
    \textbf{Lang} & \multicolumn{2}{c}{\textbf{Docs}} & \multicolumn{2}{c}{\textbf{Sents}} & \multicolumn{2}{c}{\textbf{Mean len}} \\
    \cmidrule(lr){2-3}\cmidrule(lr){4-5}\cmidrule(lr){6-7}
    & dev & test & dev & test & doc & sent \\
    \midrule
    en-fr & 1  & 1  & 37  & 447  & 22408 & 92  \\
    en-ro & 1  & 10 & 17  & 248  & 2901  & 119 \\
    en-ja & 8  & 7  & 653 & 582  & 5516  & 66  \\
    en-zh & 47 & 52 & 450 & 450  & 646   & 70  \\
    la-gr & 1  & 1  & 25  & 75   & 5837  & 116 \\
    en-cz & -- & 33 & --  & 1636 & 6318  & 126 \\
    \bottomrule
  \end{tabular}
  \caption{Word alignment dataset statistics. Lengths are target-side (\textit{text\_b}) character counts; only means shown for space. \texttt{--} indicates no data for that split.}
  \label{tab:alignment-data}
\end{table}

\subsection{Metric formulas}
\label{app:eval-formulas}
Gold alignments usually consist of two sets: sure alignments $S$ and possible alignments $P$.

\begin{equation}
\text{Precision}(\mathcal{A}, \mathcal{P}) = \frac{|\mathcal{A} \cap \mathcal{P}|}{|\mathcal{A}|}
\end{equation}

\begin{equation}
\text{Recall}(\mathcal{A}, \mathcal{S}) = \frac{|\mathcal{A} \cap \mathcal{S}|}{|\mathcal{S}|}
\end{equation}

\begin{equation}
F_1 = \frac{2 \cdot Precision \cdot Recall}{Precision + Recall}
\end{equation}

\begin{equation}
\text{AER}(\mathcal{A}, \mathcal{S}, \mathcal{P}) = 1 - \frac{|\mathcal{A} \cap \mathcal{S}| + |\mathcal{A} \cap \mathcal{P}|}{|\mathcal{A}| + |\mathcal{S}|}
\end{equation}

\noindent{}Where $\mathcal{P}$ is defined as $S \cup P$.
Metrics are reported per language pair.

\section{Details on Omission and Addition Errors in WMT24 System Outputs}
\label{app:wmt}

We evaluate our experiments for translation coverage evaluation with human annotated WMT24 system translations~\cite{Kocmi2024-fw, Freitag2024-vz}. In Table~\ref{tab:wmt-overview}, we provide an overview over the total number of documents included in the dataset for language pair, the number of documents, which include at least on addition or omission, information about length and total error percentage.

\begin{table}[h]
  \centering\small
  \setlength{\tabcolsep}{4pt}
  \begin{tabular}{l rr rr rr rr}
    \toprule
    \textbf{Lang} & \multicolumn{2}{c}{\textbf{Docs}} & \multicolumn{2}{c}{\textbf{Sents}} & \multicolumn{2}{c}{\textbf{Mean len}} & \multicolumn{2}{c}{\textbf{Err \%}} \\
    \cmidrule(lr){2-3}\cmidrule(lr){4-5}\cmidrule(lr){6-7}\cmidrule(lr){8-9}
    & tot. & err. & tot. & err. & doc & sent & add. & omi. \\
    \midrule
    en-de & 3230 & 284 & 9232 & 320 & 724 & 253 & 0.3 & 0.2 \\
    ja-zh & 2940 & 554 & 8385 & 617 & 264 & 92  & 1.0 & 1.1 \\
    \bottomrule
  \end{tabular}
  \caption{WMT 2024 MQM dataset statistics. Doc/sent lengths are target-side character counts; only means shown. \textit{err.} columns count documents/sentences containing at least one omission or addition. \textit{add.} and \textit{omi.} report the fraction (\%) of non-punctuation characters labeled as additions or omissions.}
  \label{tab:wmt-overview}
\end{table}

\section{LLM Baseline}
\label{app:llm-baseline}
The 1-shot prompt is shown in Figure~\ref{fig:prompt}. The example in the prompt is a random text pair with gold labels of the corresponding language pair from development set. If the language pair does not have a development set, the example is from en--fr.

\begin{figure*}[t]
\begin{tcolorbox}[promptbox]
\textbf{[SYSTEM]}\\
You are a helpful AI assistant for word alignment.\\[4pt]

\textbf{[USER]}\\
Please give me the word alignment information between the following two texts. The parallel texts:\\
each\_0 of\_1 them\_2 is\_3 very\_4 complex\_5 ,\_6 but\_7 the\_8 link\_9 between\_10 the\_11 two\_12 is\_13 even\_14 more\_15 complex\_16 which\_17 makes\_18 the\_19 whole\_20 situation\_21 for\_22 most\_23 people\_24 understandably\_25 confusing\_26 .\_27\\[4pt]
chacun\_0 en\_1 lui\_2 -\_3 m\^{e}me\_4 est\_5 tr\`{e}s\_6 complexe\_7 et\_8 le\_9 lien\_10 entre\_11 les\_12 deux\_13 le\_14 est\_15 encore\_16 davantage\_17 de\_18 sorte\_19 que\_20 pour\_21 beaucoup\_22 la\_23 situation\_24 pr\'{e}sente\_25 est\_26 confuse\_27 .\_28\\[4pt]

\textbf{[ASSISTANT]}\\
\{"alignments": [\{"source\_idx": 0, "target\_idxs": [0]\}, \{"source\_idx": 3, "target\_idxs": [5]\}, \{"source\_idx": 4, "target\_idxs": [6]\}, \{"source\_idx": 5, "target\_idxs": [7]\}, \{"source\_idx": 8, "target\_idxs": [9]\}, \{"source\_idx": 9, "target\_idxs": [10]\}, \{"source\_idx": 10, "target\_idxs": [11]\}, \{"source\_idx": 11, "target\_idxs": [12]\}, \{"source\_idx": 12, "target\_idxs": [13]\}, \{"source\_idx": 13, "target\_idxs": [15]\}, \{"source\_idx": 19, "target\_idxs": [23]\}, \{"source\_idx": 21, "target\_idxs": [24]\}, \{"source\_idx": 22, "target\_idxs": [21]\}, \{"source\_idx": 27, "target\_idxs": [28]\}]\}\\[4pt]

\textbf{[USER]}\\
Please give me the word alignment information between the following two texts. The parallel texts: [...]\\[4pt]

\textbf{[ASSISTANT]}\\
\{"alignments": [...]\}
\end{tcolorbox}
\caption{1-shot prompt example used for the LLM baseline based on the ``full mode'' prompt introduced by~\citet{Miao2025-sk}.}
\label{fig:prompt}
\end{figure*}

\section{Experiments on the Development Set}
\label{app:dev}

\subsection{Model selection and granularity gap}

Table~\ref{tab:pilot-results-avg} reports pilot experiment results on the development set across a larger set of embedding models (including \texttt{XLM-R} (large; ~\citep{Conneau2020-yl} and \texttt{mBERT}~\citep{Devlin2019-nj} at the sentence-level as well as \texttt{EuroBERT-610m}~\citep{Boizard2025-ye} and \texttt{Qwen3-Embedding-0.6}~\citep{Zhang2025-kx} for both granularities and both \SimAlign{} variants as macro-averages across languages. The results reveal a degradation when moving from sentence- to document-level alignment. They also motivate our model selection for the main experiments: \texttt{LaBSE},  \texttt{Qwen-Embedding-4B}, and \texttt{mmBERT-base} emerge as the strongest multilingual embedding models overall.

\begin{table}[t]
\centering
\small
\begin{tabular}{lrrrr}
\hline
\textbf{Model / Method} & \textbf{P} & \textbf{R} & \textbf{F1} & \textbf{AER} \\
\hline

\multicolumn{5}{l}{\textsc{Sentence-level}} \\

\multicolumn{5}{l}{\hspace{1em}\textit{XLM-R-large}} \\
\SimAlign{} + Argmax  & 0.777 & 0.635 & 0.692 & 0.308 \\
\SimAlign{} + Itermax & 0.689 & 0.692 & 0.685 & 0.317 \\

\multicolumn{5}{l}{\hspace{1em}\textit{mBERT}} \\
\SimAlign{} + Argmax  & 0.816 & 0.625 & 0.701 & 0.297 \\
\SimAlign{} + Itermax & 0.711 & 0.711 & 0.707 & 0.295 \\

\multicolumn{5}{l}{\hspace{1em}\textit{mmBERT}} \\
\SimAlign{} + Argmax  & 0.844 & 0.710 & 0.763 & 0.236 \\
\SimAlign{} + Itermax & 0.733 & 0.791 & 0.755 & 0.248 \\

\multicolumn{5}{l}{\hspace{1em}\textit{EuroBERT-610m}} \\
\SimAlign{} + Argmax  & 0.671 & 0.516 & 0.573 & 0.426 \\
\SimAlign{} + Itermax & 0.545 & 0.596 & 0.563 & 0.442 \\

\multicolumn{5}{l}{\hspace{1em}\textit{LaBSE}} \\
\SimAlign{} + Argmax  & \underline{0.858} & 0.727 & \underline{0.781} & \underline{0.218} \\
\SimAlign{} + Itermax & 0.753 & \underline{0.799} & 0.772 & 0.232 \\

\multicolumn{5}{l}{\hspace{1em}\textit{Qwen-0.6B}} \\
\SimAlign{} + Argmax  & 0.780 & 0.601 & 0.670 & 0.329 \\
\SimAlign{} + Itermax & 0.672 & 0.681 & 0.669 & 0.334 \\

\multicolumn{5}{l}{\hspace{1em}\textit{Qwen-4B}} \\
\SimAlign{} + Argmax  & 0.807 & 0.638 & 0.702 & 0.297 \\
\SimAlign{} + Itermax & 0.699 & 0.682 & 0.685 & 0.316 \\

\hline

\multicolumn{5}{l}{\textsc{Document-level}} \\

\multicolumn{5}{l}{\hspace{1em}\textit{mmBERT}} \\
\SimAlign{} + Argmax  & 0.706 & 0.453 & 0.532 & 0.467 \\
\SimAlign{} + Itermax & 0.616 & 0.531 & 0.553 & 0.449 \\

\multicolumn{5}{l}{\hspace{1em}\textit{EuroBERT-610m}} \\
\SimAlign{} + Argmax  & 0.621 & 0.358 & 0.427 & 0.572 \\
\SimAlign{} + Itermax & 0.508 & 0.414 & 0.433 & 0.571 \\

\multicolumn{5}{l}{\hspace{1em}\textit{LaBSE-concatenated-sentences}} \\
\SimAlign{} + Argmax  & \textbf{0.771} & 0.495 & 0.594 & 0.401 \\
\SimAlign{} + Itermax & 0.688 & 0.574 & 0.617 & 0.378 \\

\multicolumn{5}{l}{\hspace{1em}\textit{Qwen-0.6B}} \\
\SimAlign{} + Argmax  & 0.740 & 0.496 & 0.580 & 0.419 \\
\SimAlign{} + Itermax & 0.640 & 0.565 & 0.588 & 0.413 \\

\multicolumn{5}{l}{\hspace{1em}\textit{Qwen-4B}} \\
\SimAlign{} + Argmax  & 0.768 & 0.568 & \textbf{0.643} & \textbf{0.355} \\
\SimAlign{} + Itermax & 0.657 & \textbf{0.642} & 0.639 & 0.363 \\

\hline
\end{tabular}
\caption{Macro-average pilot word alignment development results across language pairs. Best sentence-level results are \underline{underlined} and best document-level results are \textbf{bolded}. For AER lower is better.}
\label{tab:pilot-results-avg}
\end{table}

\subsection{Hyperparameter selection}
\label{app:hyp-selection}
\paragraph{Model layer} Previous work indicates that word alignment performance varies with embedding model layers and that the default last layer oftentimes underperforms in comparison to middle layers~\citep{Zhang2019-fm, Jalili-Sabet2020-rb}. Hence, we search for the best layer for each language pair and granularity. 

\begin{figure}[h]
  \centering

  \resizebox{0.9\columnwidth}{!}{%
    \includegraphics{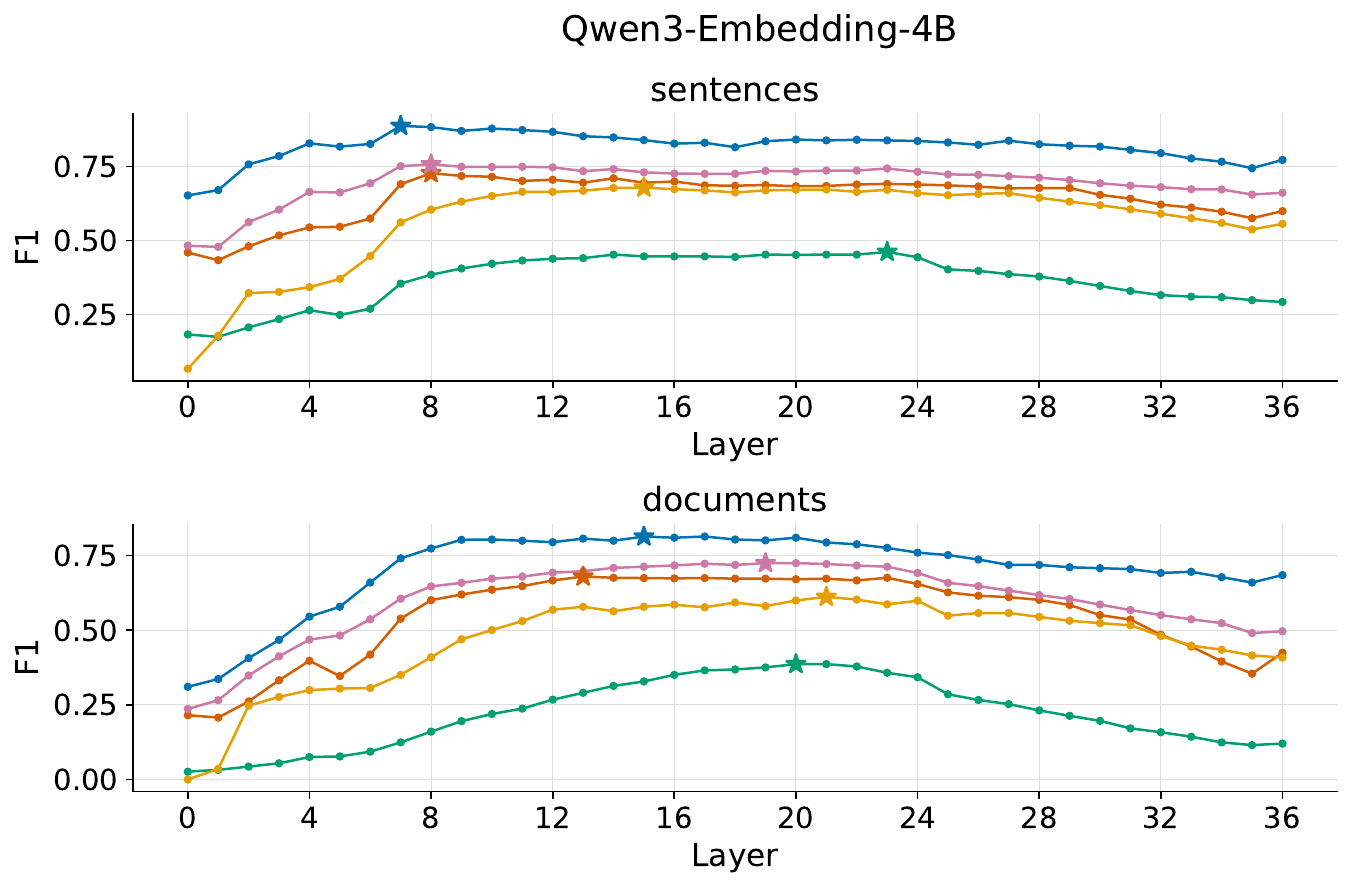}
  }

  \vspace{2mm}

  \resizebox{0.9\columnwidth}{!}{%
    \includegraphics{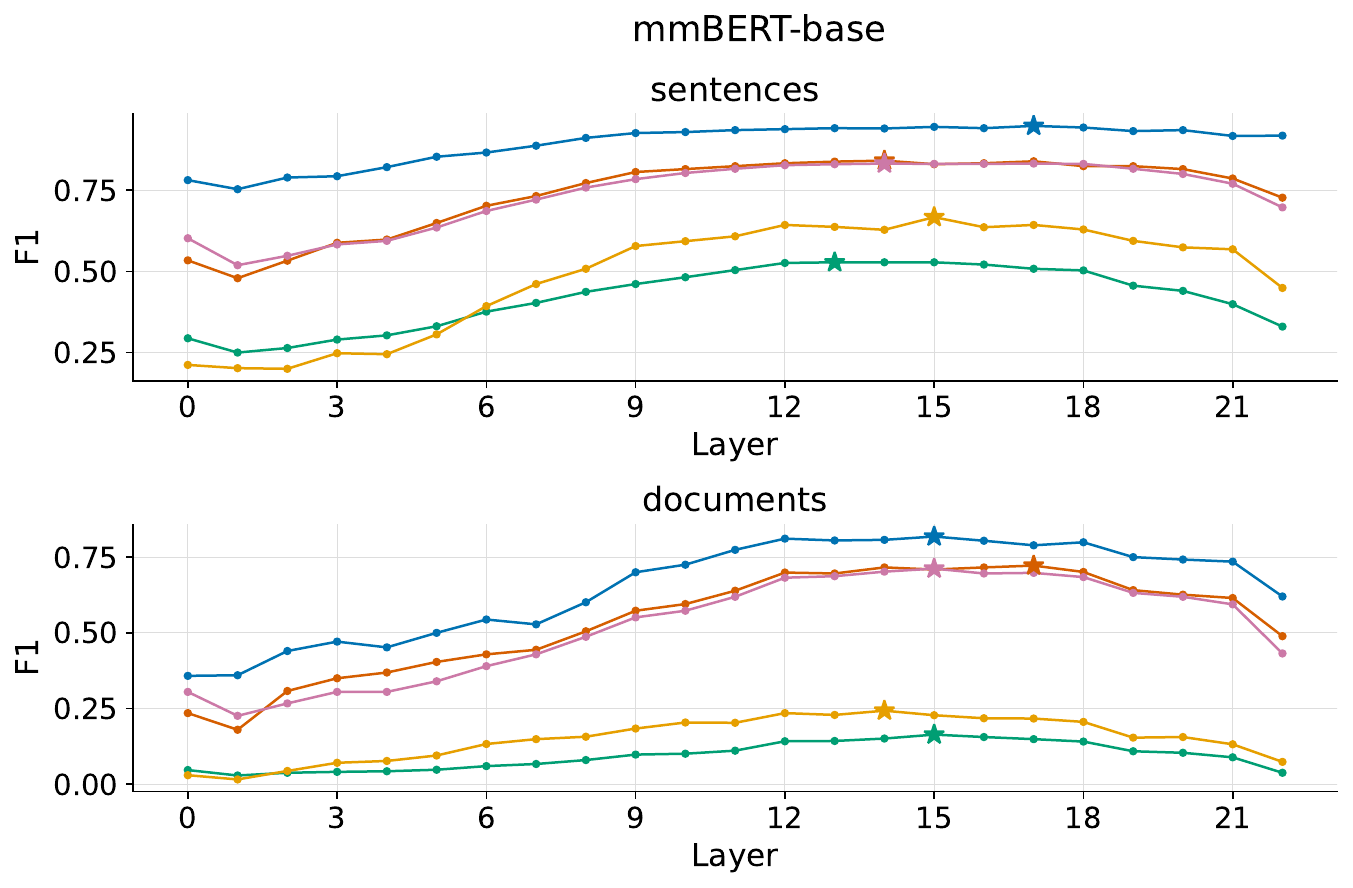}
  }

  \vspace{2mm}

  \resizebox{0.9\columnwidth}{!}{%
    \includegraphics{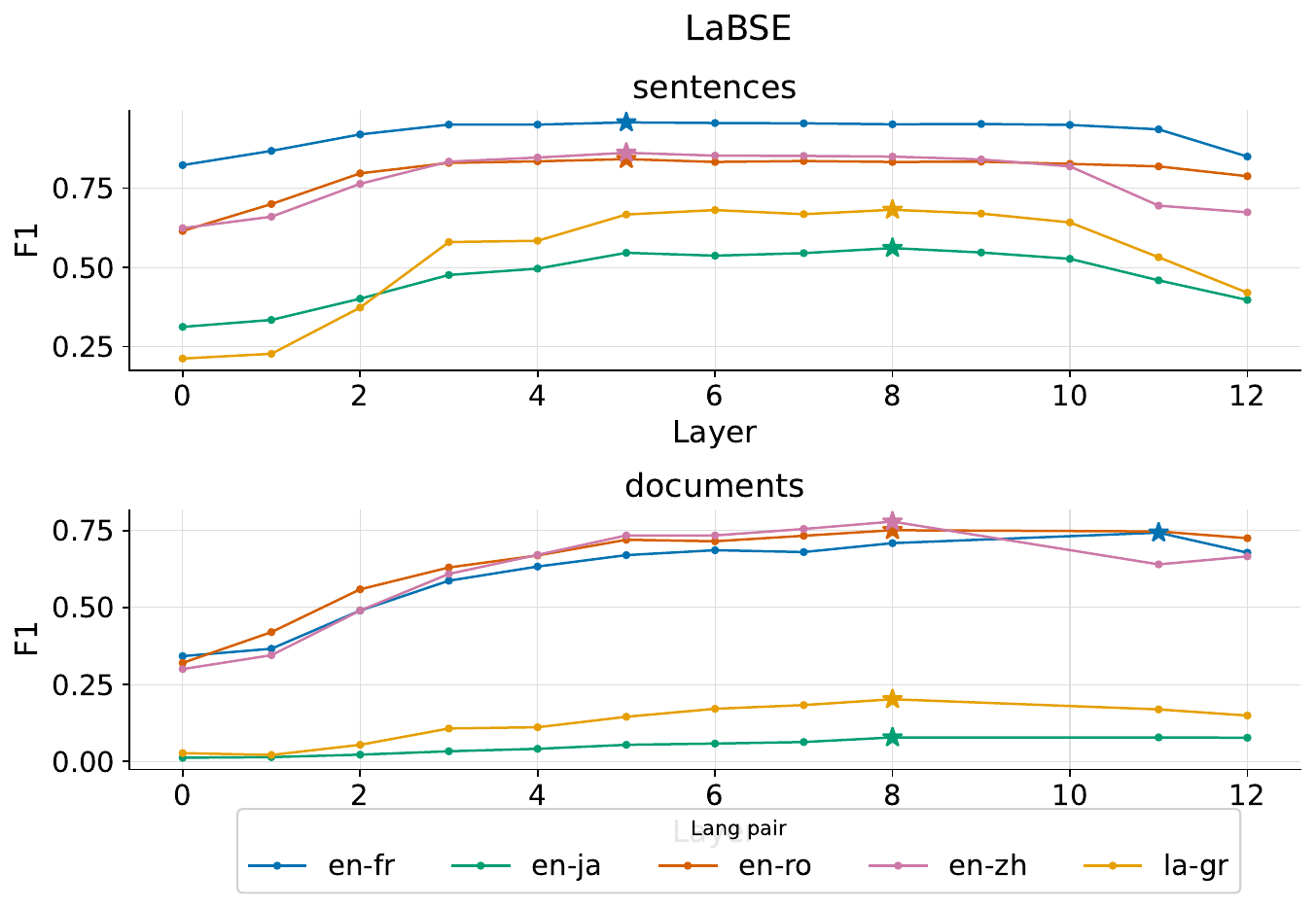}
  }

  \caption{F1 word alignment results across layers for different language pairs with Argmax on the development set for both granularities for \texttt{Qwen3-Embedding-4B}, \texttt{mmBERT-base}, and \texttt{LaBSE-concatenated-sentences}.}
  \label{fig:layer-results}
\end{figure}

The layer search reported in Figure~\ref{fig:layer-results} reveals that the layer selection becomes more important at the document level with differences of over 25 F1 points between the best layer and the last. Furthermore, document pairs tend to benefit from later layers than sentence pairs for word alignment. For all further reported results we use the best performing layers per language pair and granularity (marked with a star in Figure~\ref{fig:layer-results}).

\begin{figure}[h]
  \centering

  \resizebox{\columnwidth}{!}{%
    \includegraphics{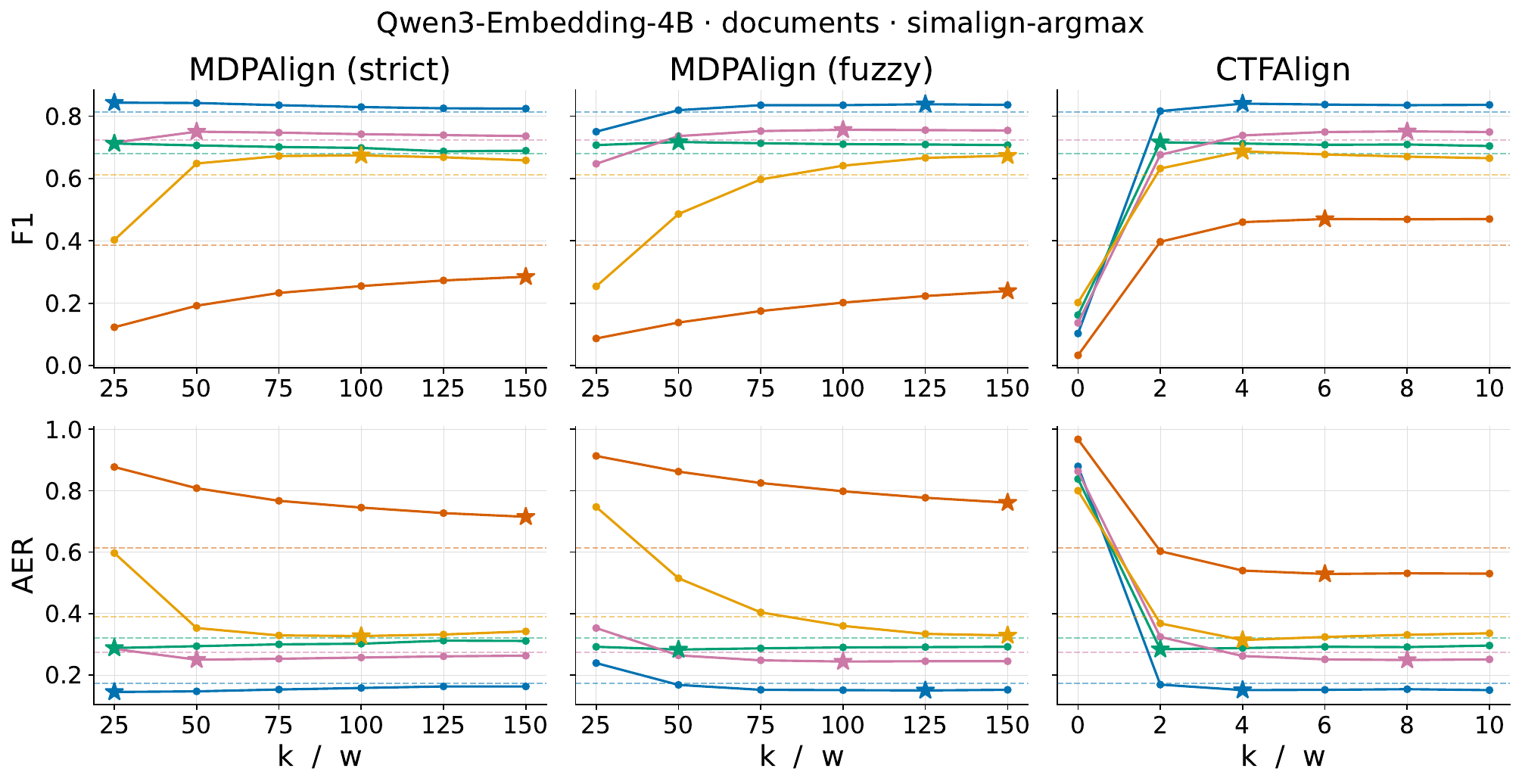}
  }

  \vspace{2mm}

  \resizebox{\columnwidth}{!}{%
    \includegraphics{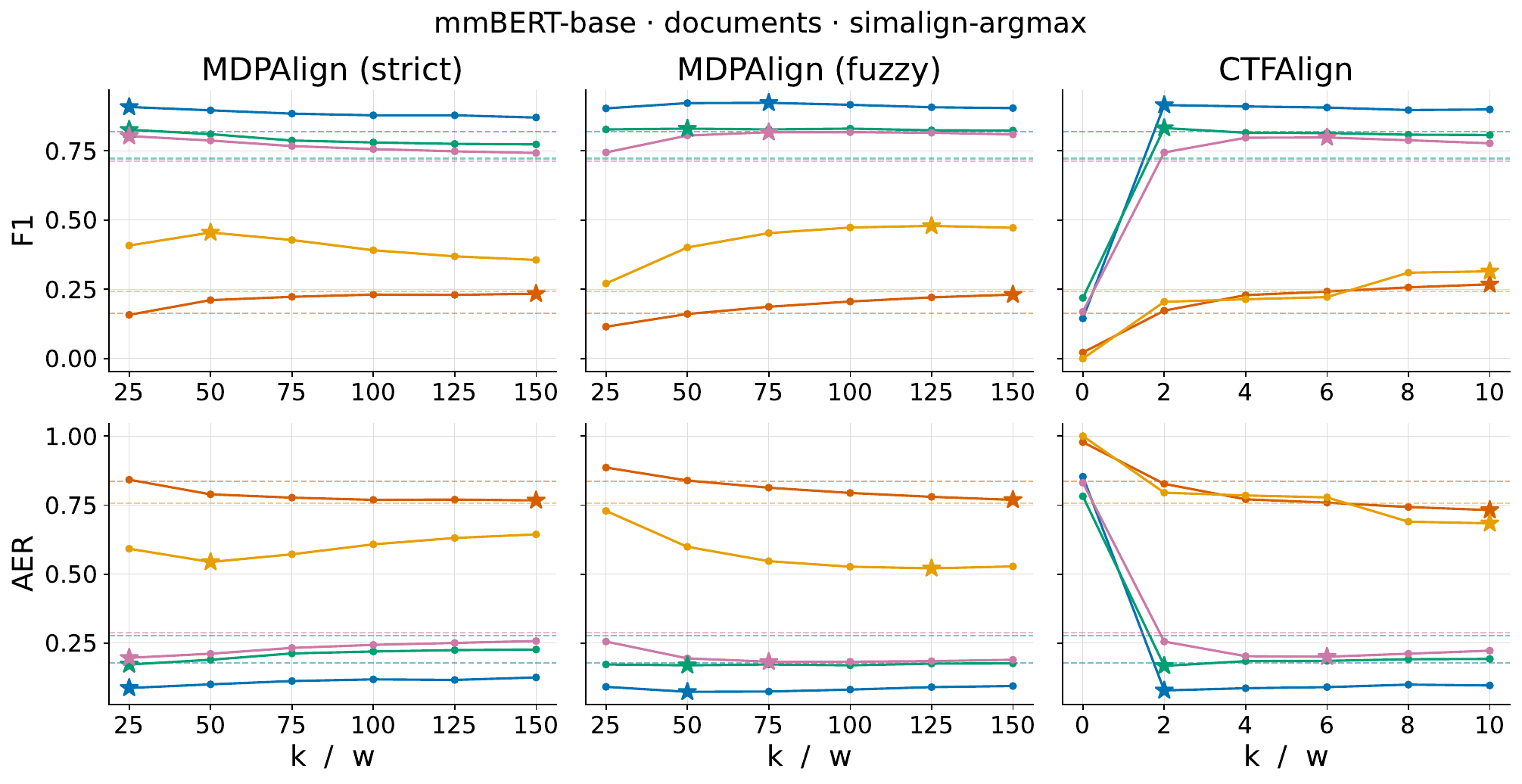}
  }

  \vspace{2mm}

  \resizebox{\columnwidth}{!}{%
    \includegraphics{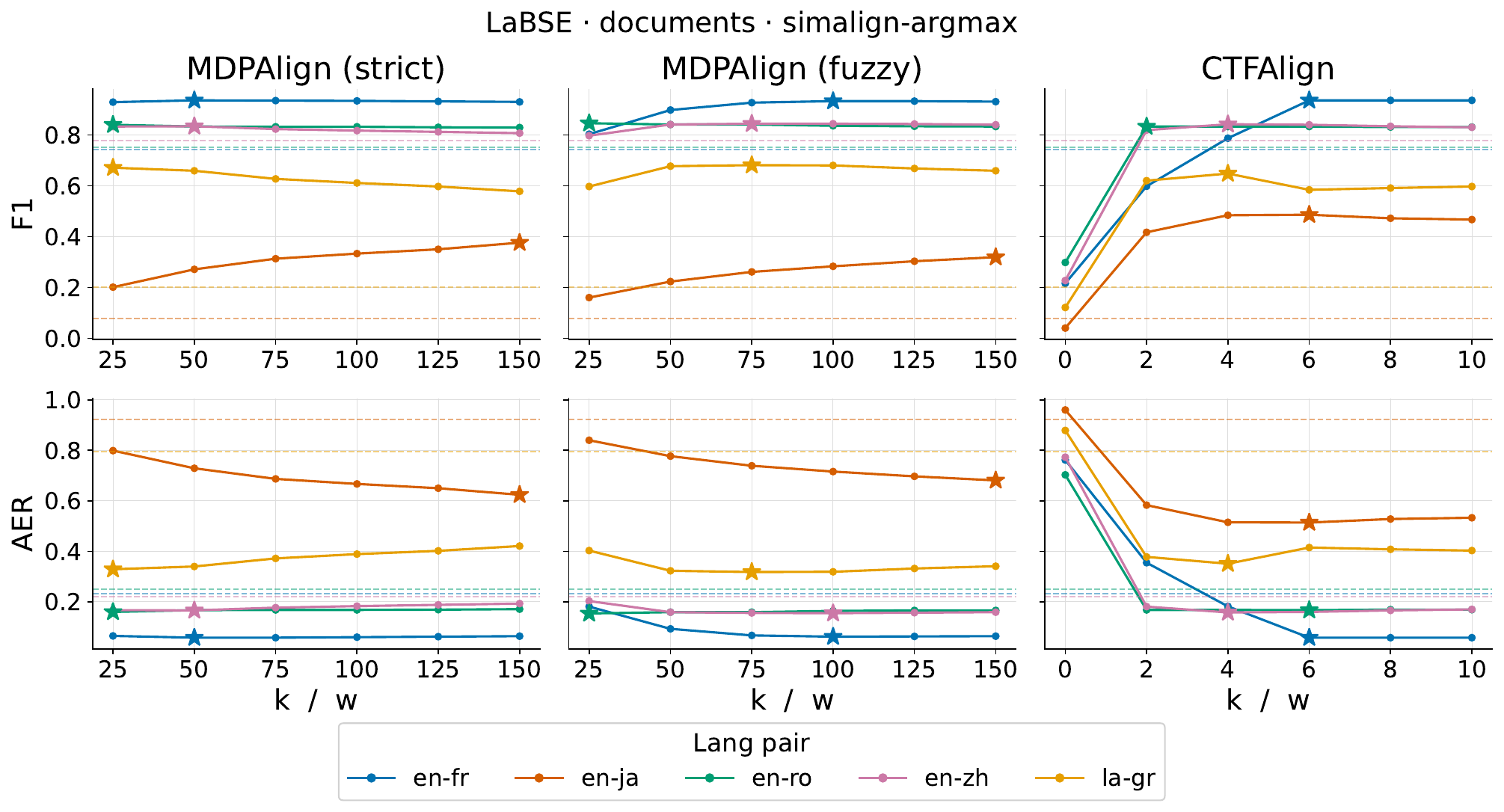}
  }

  \caption{AER (lower is better) and F1 (higher is better) results for different $k$/$w$ values for \texttt{Qwen3-Embedding-4B}, \texttt{mmBERT-base}, and \texttt{LaBSE-concatenated-sentences}.}
  \label{fig:k-comb}
\end{figure}

\paragraph{Width hyperparameters $k$ and $w$} Both hyperparameters first increase performance with a growing value for their corresponding algorithm but do not substantially change the performance once a plateau is reached (as shown in Figure~\ref{fig:k-comb}). Hence, we decide to fix the bandwidth parameter $k$ and $w$ at the first value that reaches that plateau on average across language pairs and models (Table~\ref{tab:k-search}). For \MDPAlign{} (strict) a narrow bandwidth of $k=50$ is optimal, where the maximum value of $k=150$ for \MDPAlign{} (fuzzy) and $w=8$ for \CTFAlign{} performs best. 
For all further experiments, we fix the values according to Table~\ref{tab:k-search}.

\begin{table}[t]
\centering
\resizebox{\columnwidth}{!}{
\begin{tabular}{lrrr}
\hline
\textbf{Method} & \textbf{en--de} & \textbf{en--fr} & \textbf{en--it} \\
\hline

\multicolumn{4}{l}{\textsc{Last layer}} \\
\hspace{1em}DiffAlign & 0.186 & 0.098 & 0.206 \\
\hspace{1em}DiffAlign\textsubscript{MDP+Argmax} & 0.227 & 0.110 & 0.228 \\
\hspace{1em}DiffAlign\textsubscript{CTF+Itermax} & 0.262 & 0.140 & 0.254 \\

\hline

\multicolumn{4}{l}{\textsc{Layer 20}} \\
\hspace{1em}DiffAlign & 0.286 & 0.156 & 0.318 \\
\hspace{1em}DiffAlign\textsubscript{MDP+Argmax} & 0.301 & 0.168 & 0.309 \\
\hspace{1em}DiffAlign\textsubscript{CTF+Argmax} & \textbf{0.346} & \textbf{0.208} & \textbf{0.365} \\
\hspace{1em}DiffAlign\textsubscript{CTF+Itermax} & 0.332 & 0.199 & 0.351 \\

\hline
\end{tabular}
}
\caption{SwissGov-RSD development set Spearman $\rho$ for DiffAlign variants using \texttt{Qwen3-Embedding-4B} across en--de, en--fr, and en--it. Bold indicates the best result per language pair.}
\label{tab:diffalign-spearman-dev}
\end{table}

\begin{table}[t]
\centering\small
\begin{tabular}{lr rr}
\hline
\textbf{Approach} & \textbf{$k$/$w$} & \textbf{F1} & \textbf{AER} \\
\hline
\multirow{6}{*}{CTFAlign} & 0 & 0.140 & 0.858 \\
 & 2 & 0.626 & 0.369 \\
 & 4 & 0.666 & 0.331 \\
 & 6 & 0.673 & 0.325 \\
 & 8 & \textbf{0.677} & \textbf{0.321} \\
 & 10 & \textbf{0.677} & \textbf{0.322} \\
\hline
\multirow{6}{*}{\MDPAlign{} (strict)} & 25 & 0.625 & 0.374 \\
 & 50 & \textbf{0.655} & \textbf{0.343} \\
 & 75 & 0.654 & 0.345 \\
 & 100 & 0.651 & 0.348 \\
 & 125 & 0.648 & 0.351 \\
 & 150 & 0.646 & 0.353 \\
\hline
\multirow{6}{*}{\MDPAlign{} (fuzzy)} & 25 & 0.567 & 0.431 \\
 & 50 & 0.633 & 0.365 \\
 & 75 & 0.656 & 0.343 \\
 & 100 & 0.664 & 0.334 \\
 & 125 & 0.668 & 0.331 \\
 & 150 & \textbf{0.669} & \textbf{0.330} \\
\hline
\end{tabular}
\caption{Average dev-set F1 and AER (lower is better) across \texttt{LaBSE}, \texttt{mmBERT}, and \texttt{Qwen} for each hyperparameter value. Approaches were combined with Argmax. Best per approach and column in \textbf{bold}.}
\label{tab:k-search}
\end{table}

\subsection{Results \& Discussion}
\label{app:dev-res}

\begin{table*}[t]
\resizebox{\textwidth}{!}{
\begin{tabular}{l @{\hspace{8pt}} rrrr @{\hspace{8pt}} rrrr @{\hspace{8pt}} rrrr @{\hspace{8pt}} rrrr @{\hspace{8pt}} rrrr @{\hspace{8pt}}| rrrr}
\hline
\textbf{Model} & \multicolumn{4}{c}{\textbf{en-fr}} & \multicolumn{4}{c}{\textbf{en-ro}} & \multicolumn{4}{c}{\textbf{en-ja}} & \multicolumn{4}{c}{\textbf{en-zh}} & \multicolumn{4}{c}{\textbf{la-gr}} & \multicolumn{4}{c}{\textbf{Avg}} \\
& \textbf{P} & \textbf{R} & \textbf{F1} & \textbf{AER} & \textbf{P} & \textbf{R} & \textbf{F1} & \textbf{AER} & \textbf{P} & \textbf{R} & \textbf{F1} & \textbf{AER} & \textbf{P} & \textbf{R} & \textbf{F1} & \textbf{AER} & \textbf{P} & \textbf{R} & \textbf{F1} & \textbf{AER} & \textbf{P} & \textbf{R} & \textbf{F1} & \textbf{AER} \\
\hline
\multicolumn{25}{l}{\textsc{Sentence-level}} \\
\textit{GPT-5.4-mini} & 0.820 & 0.799 & 0.809 & 0.187 & 0.632 & 0.592 & 0.611 & 0.389 & 0.255 & \underline{0.576} & 0.354 & 0.646 & 0.547 & 0.669 & 0.602 & 0.398 & \underline{0.762} & \underline{0.849} & \underline{0.803} & \underline{0.198} & 0.603 & 0.697 & 0.636 & 0.364 \\
\multicolumn{25}{l}{\hspace{1em}\textit{LaBSE}} \\
\SimAlign{} + Argmax & 0.966 & 0.950 & \underline{0.958} & \underline{0.040} & \underline{0.971} & 0.743 & 0.842 & 0.158 & \underline{0.761} & 0.444 & 0.561 & 0.439 & \underline{0.892} & 0.834 & \underline{0.862} & \underline{0.138} & 0.702 & 0.664 & 0.682 & 0.317 & \underline{0.858} & 0.727 & \underline{0.781} & \underline{0.218} \\
\SimAlign{} + Itermax & 0.886 & \underline{0.979} & 0.930 & 0.083 & 0.874 & 0.834 & 0.854 & 0.146 & 0.652 & 0.532 & \underline{0.586} & \underline{0.414} & 0.777 & \underline{0.896} & 0.832 & 0.168 & 0.604 & 0.753 & 0.670 & 0.331 & 0.759 & \underline{0.799} & 0.774 & 0.228 \\
\multicolumn{25}{l}{\hspace{1em}\textit{mmBERT}} \\
\SimAlign{} + Argmax & \underline{0.968} & 0.929 & 0.948 & 0.048 & 0.960 & 0.749 & 0.841 & 0.159 & 0.750 & 0.408 & 0.528 & 0.471 & 0.882 & 0.788 & 0.832 & 0.167 & 0.660 & 0.674 & 0.667 & 0.333 & 0.844 & 0.710 & 0.763 & 0.236 \\
\SimAlign{} + Itermax & 0.868 & 0.967 & 0.915 & 0.098 & 0.875 & \underline{0.842} & \underline{0.858} & \underline{0.142} & 0.635 & 0.502 & 0.561 & 0.439 & 0.746 & 0.883 & 0.809 & 0.192 & 0.539 & 0.760 & 0.631 & 0.371 & 0.733 & 0.791 & 0.755 & 0.248 \\
\multicolumn{25}{l}{\hspace{1em}\textit{Qwen}} \\
\SimAlign{} + Argmax & 0.943 & 0.837 & 0.887 & 0.104 & 0.955 & 0.587 & 0.727 & 0.273 & 0.663 & 0.353 & 0.461 & 0.540 & 0.837 & 0.691 & 0.757 & 0.243 & 0.638 & 0.723 & 0.678 & 0.323 & 0.807 & 0.638 & 0.702 & 0.297 \\
\SimAlign{} + Itermax & 0.824 & 0.891 & 0.856 & 0.151 & 0.828 & 0.648 & 0.727 & 0.273 & 0.548 & 0.434 & 0.484 & 0.515 & 0.737 & 0.738 & 0.737 & 0.262 & 0.557 & 0.699 & 0.620 & 0.381 & 0.699 & 0.682 & 0.685 & 0.316 \\
\\
\hline
\multicolumn{25}{l}{\textsc{Document-level}} \\
\textit{GPT-5.4-mini} & 0.072 & 0.068 & 0.070 & 0.929 & 0.024 & 0.022 & 0.023 & 0.977 & 0.010 & 0.010 & 0.010 & 0.990 & 0.065 & 0.079 & 0.071 & 0.929 & 0.095 & 0.114 & 0.104 & 0.897 & 0.053 & 0.059 & 0.056 & 0.944 \\
\multicolumn{25}{l}{\hspace{1em}\textit{LaBSE}} \\
\SimAlign{} + Argmax & 0.933 & 0.533 & 0.678 & 0.297 & 0.950 & 0.621 & 0.751 & 0.249 & 0.632 & 0.236 & 0.344 & 0.656 & 0.858 & 0.714 & 0.779 & 0.220 & 0.480 & 0.370 & 0.418 & 0.581 & 0.771 & 0.495 & 0.594 & 0.401 \\
\MDPAlign{} + Argmax (strict) & \textbf{0.969} & 0.905 & \textbf{0.936} & \textbf{0.058} & \textbf{0.976} & 0.727 & 0.833 & 0.167 & 0.451 & 0.194 & 0.271 & 0.729 & 0.868 & 0.802 & 0.834 & 0.166 & \textbf{0.685} & 0.635 & 0.659 & 0.340 & 0.790 & 0.653 & 0.707 & 0.292 \\
\MDPAlign{} + Argmax (fuzzy) & 0.958 & 0.905 & 0.931 & 0.064 & 0.971 & 0.730 & 0.833 & 0.166 & 0.514 & 0.231 & 0.319 & 0.681 & \textbf{0.871} & 0.812 & \textbf{0.840} & \textbf{0.159} & 0.684 & 0.635 & 0.659 & 0.341 & 0.800 & 0.663 & 0.716 & 0.282 \\
CTFAlign + Argmax & \textbf{0.969} & 0.905 & \textbf{0.936} & \textbf{0.058} & 0.955 & 0.736 & 0.831 & 0.169 & \textbf{0.722} & 0.351 & 0.472 & 0.528 & 0.867 & 0.804 & 0.834 & 0.166 & 0.645 & 0.546 & 0.591 & 0.408 & \textbf{0.832} & 0.668 & \textbf{0.733} & \textbf{0.266} \\
\MDPAlign{} + Itermax (strict) & 0.910 & 0.929 & 0.919 & 0.083 & 0.868 & 0.829 & 0.848 & 0.152 & 0.363 & 0.238 & 0.288 & 0.713 & 0.758 & 0.878 & 0.814 & 0.187 & 0.580 & 0.721 & 0.643 & 0.359 & 0.696 & 0.719 & 0.702 & 0.299 \\
\MDPAlign{} + Itermax (fuzzy) & 0.906 & 0.935 & 0.920 & 0.083 & 0.870 & \textbf{0.831} & \textbf{0.850} & \textbf{0.150} & 0.424 & 0.289 & 0.344 & 0.656 & 0.760 & \textbf{0.883} & 0.817 & 0.184 & 0.583 & 0.721 & 0.645 & 0.357 & 0.709 & \textbf{0.732} & 0.715 & 0.286 \\
CTFAlign + Itermax & 0.909 & 0.929 & 0.919 & 0.084 & 0.872 & 0.829 & \textbf{0.850} & \textbf{0.150} & 0.608 & \textbf{0.462} & \textbf{0.525} & \textbf{0.475} & 0.757 & 0.874 & 0.811 & 0.189 & 0.512 & 0.506 & 0.509 & 0.491 & 0.732 & 0.720 & 0.723 & 0.278 \\
\multicolumn{25}{l}{\hspace{1em}\textit{mmBERT}} \\
\SimAlign{} + Argmax & 0.851 & 0.787 & 0.818 & 0.178 & 0.958 & 0.579 & 0.722 & 0.278 & 0.533 & 0.097 & 0.164 & 0.836 & 0.862 & 0.606 & 0.712 & 0.288 & 0.324 & 0.195 & 0.243 & 0.756 & 0.706 & 0.453 & 0.532 & 0.467 \\
\MDPAlign{} + Argmax (strict) & 0.917 & 0.876 & 0.896 & 0.101 & 0.953 & 0.705 & 0.810 & 0.190 & 0.372 & 0.147 & 0.211 & 0.789 & 0.857 & 0.728 & 0.787 & 0.212 & 0.467 & 0.444 & 0.455 & 0.544 & 0.713 & 0.580 & 0.632 & 0.367 \\
\MDPAlign{} + Argmax (fuzzy) & 0.909 & 0.899 & 0.904 & 0.095 & 0.961 & 0.719 & 0.823 & 0.177 & 0.401 & 0.162 & 0.231 & 0.769 & 0.867 & 0.759 & 0.809 & 0.190 & 0.486 & 0.459 & 0.472 & 0.528 & 0.725 & 0.600 & 0.648 & 0.352 \\
CTFAlign + Argmax & 0.916 & 0.879 & 0.897 & 0.100 & 0.958 & 0.699 & 0.808 & 0.192 & 0.495 & 0.174 & 0.257 & 0.743 & 0.855 & 0.731 & 0.788 & 0.212 & 0.345 & 0.281 & 0.310 & 0.690 & 0.714 & 0.553 & 0.612 & 0.387 \\
\MDPAlign{} + Itermax (strict) & 0.776 & 0.947 & 0.853 & 0.165 & 0.874 & 0.798 & 0.834 & 0.166 & 0.296 & 0.187 & 0.229 & 0.771 & 0.743 & 0.846 & 0.791 & 0.209 & 0.364 & 0.548 & 0.437 & 0.564 & 0.611 & 0.665 & 0.629 & 0.375 \\
\MDPAlign{} + Itermax (fuzzy) & 0.771 & \textbf{0.956} & 0.854 & 0.166 & 0.881 & 0.811 & 0.845 & 0.156 & 0.323 & 0.206 & 0.252 & 0.748 & 0.747 & 0.863 & 0.801 & 0.200 & 0.380 & 0.565 & 0.454 & 0.546 & 0.620 & 0.680 & 0.641 & 0.363 \\
CTFAlign + Itermax & 0.782 & 0.947 & 0.857 & 0.159 & 0.866 & 0.787 & 0.825 & 0.176 & 0.411 & 0.225 & 0.291 & 0.709 & 0.745 & 0.836 & 0.788 & 0.212 & 0.251 & 0.341 & 0.289 & 0.711 & 0.611 & 0.627 & 0.610 & 0.393 \\
\multicolumn{25}{l}{\hspace{1em}\textit{mmBERT-concatenated-sentences}} \\
\SimAlign{} + Argmax & 0.894 & 0.781 & 0.834 & 0.159 & 0.948 & 0.563 & 0.706 & 0.294 & 0.656 & 0.140 & 0.231 & 0.770 & 0.868 & 0.568 & 0.687 & 0.313 & 0.439 & 0.299 & 0.356 & 0.644 & 0.761 & 0.470 & 0.563 & 0.436 \\
\MDPAlign{} + Argmax (strict) & 0.947 & 0.891 & 0.918 & 0.077 & 0.967 & 0.736 & 0.836 & 0.164 & 0.296 & 0.118 & 0.169 & 0.832 & 0.870 & 0.756 & 0.809 & 0.191 & 0.506 & 0.521 & 0.513 & 0.487 & 0.717 & 0.604 & 0.649 & 0.350 \\
\MDPAlign{} + Argmax (fuzzy) & 0.946 & 0.899 & 0.922 & 0.074 & 0.973 & 0.734 & 0.837 & 0.163 & 0.437 & 0.173 & 0.248 & 0.752 & 0.884 & 0.756 & 0.815 & 0.185 & 0.601 & 0.607 & 0.604 & 0.396 & 0.768 & 0.634 & 0.685 & 0.314 \\
CTFAlign + Argmax & 0.937 & 0.858 & 0.896 & 0.097 & 0.973 & 0.714 & 0.824 & 0.176 & 0.524 & 0.186 & 0.275 & 0.726 & 0.867 & 0.703 & 0.776 & 0.223 & 0.584 & 0.558 & 0.571 & 0.429 & 0.777 & 0.604 & 0.668 & 0.330 \\
\MDPAlign{} + Itermax (strict) & 0.838 & 0.947 & 0.889 & 0.124 & 0.877 & 0.831 & 0.853 & 0.147 & 0.227 & 0.143 & 0.175 & 0.825 & 0.742 & 0.853 & 0.794 & 0.207 & 0.395 & 0.590 & 0.473 & 0.528 & 0.616 & 0.673 & 0.637 & 0.366 \\
\MDPAlign{} + Itermax (fuzzy) & 0.833 & 0.953 & 0.889 & 0.124 & 0.892 & 0.827 & 0.858 & 0.142 & 0.359 & 0.219 & 0.272 & 0.728 & 0.768 & 0.864 & 0.813 & 0.187 & 0.486 & 0.716 & 0.579 & 0.422 & 0.668 & 0.716 & 0.682 & 0.321 \\
CTFAlign + Itermax & 0.818 & 0.932 & 0.871 & 0.141 & 0.889 & 0.801 & 0.843 & 0.157 & 0.474 & 0.240 & 0.319 & 0.682 & 0.759 & 0.819 & 0.788 & 0.212 & 0.444 & 0.610 & 0.514 & 0.487 & 0.677 & 0.680 & 0.667 & 0.336 \\
\multicolumn{25}{l}{\hspace{1em}\textit{Qwen}} \\
\SimAlign{} + Argmax & 0.894 & 0.746 & 0.813 & 0.174 & 0.901 & 0.545 & 0.679 & 0.321 & 0.626 & 0.279 & 0.386 & 0.614 & 0.812 & 0.654 & 0.724 & 0.275 & 0.608 & 0.615 & 0.611 & 0.389 & 0.768 & 0.568 & 0.643 & 0.355 \\
\MDPAlign{} + Argmax (strict) & 0.905 & 0.787 & 0.842 & 0.147 & 0.863 & 0.597 & 0.706 & 0.294 & 0.325 & 0.136 & 0.192 & 0.808 & 0.799 & 0.706 & 0.750 & 0.250 & 0.604 & 0.699 & 0.648 & 0.353 & 0.699 & 0.585 & 0.628 & 0.370 \\
\MDPAlign{} + Argmax (fuzzy) & 0.900 & 0.781 & 0.836 & 0.152 & 0.868 & 0.597 & 0.707 & 0.292 & 0.388 & 0.173 & 0.239 & 0.761 & 0.801 & 0.713 & 0.754 & 0.245 & 0.623 & 0.731 & \textbf{0.673} & \textbf{0.329} & 0.716 & 0.599 & 0.642 & 0.356 \\
CTFAlign + Argmax & 0.896 & 0.781 & 0.835 & 0.154 & 0.868 & 0.599 & 0.709 & 0.291 & 0.639 & 0.370 & 0.469 & 0.531 & 0.798 & 0.709 & 0.751 & 0.249 & 0.622 & 0.726 & 0.670 & 0.331 & 0.765 & 0.637 & 0.687 & 0.311 \\
\MDPAlign{} + Itermax (strict) & 0.786 & 0.840 & 0.812 & 0.194 & 0.765 & 0.676 & 0.718 & 0.282 & 0.249 & 0.165 & 0.198 & 0.802 & 0.671 & 0.808 & 0.733 & 0.268 & 0.457 & 0.743 & 0.566 & 0.437 & 0.586 & 0.646 & 0.605 & 0.397 \\
\MDPAlign{} + Itermax (fuzzy) & 0.777 & 0.840 & 0.807 & 0.200 & 0.764 & 0.674 & 0.716 & 0.284 & 0.313 & 0.218 & 0.257 & 0.743 & 0.671 & 0.815 & 0.736 & 0.265 & 0.481 & 0.778 & 0.594 & 0.409 & 0.601 & 0.665 & 0.622 & 0.380 \\
CTFAlign + Itermax & 0.775 & 0.837 & 0.805 & 0.203 & 0.775 & 0.676 & 0.722 & 0.278 & 0.526 & 0.455 & 0.488 & 0.512 & 0.669 & 0.807 & 0.732 & 0.269 & 0.481 & \textbf{0.788} & 0.597 & 0.406 & 0.645 & 0.713 & 0.669 & 0.334 \\
\hline
\end{tabular}
}
\caption{Word alignment development results (P, R, F1, AER). Fixed bandwidths: \MDPAlign{} strict $k{=}50$, \MDPAlign{} fuzzy $k{=}150$, \CTFAlign{} $w{=}8$. Sentence-level rows are upper-bound baselines; best per column \underline{underlined}. Document-level best per column \textbf{bold}.}
\label{tab:alignment-results-dev}
\end{table*}

Development set results are presented in Table~\ref{tab:alignment-results-dev}. Directly applying \SimAlign{} at document level leads to degradation compared to sentence-level alignment with both Argmax and Itermax. Since both Armgax and Itermax perform comparably at the document-level, we continue to use only the argmax variants as a baseline in all further experiments. Furthermore, experiments with increasing the iteration count of Itermax leads to further degradation of the performance while also increasing runtime.
The document-level degradation is particularly severe for recall, indicating that unconstrained similarity matrices contain many competing high-scoring similarities that cause greedy alignment decisions to miss correct correspondences. Both \MDPAlign{} and \CTFAlign{} improve recall by restricting the search space and suppressing spurious long-range matches. 
Strict \MDPAlign{} has the overall lowest performance, hence we do not proceed to evaluate it on the test set in the main paper.

\section{SwissGov-RSD Development Set Results}
\label{app:swissgov}

Table~\ref{tab:diffalign-spearman-dev} shows development set results on SwissGov-RSD using \texttt{Qwen3-Embedding-4B}. Across all language pairs, applying structural constraints consistently improves over the unconstrained DiffAlign baseline, indicating that restricting the search space is beneficial for document-level semantic difference detection. 

As is the case in test experiments, our approaches outperform the baseline across all language pairs while DiffAlign\textsubscript{CTF+Argmax} achieves the strongest overall performance. The gains are especially pronounced at layer 20, which is on average across languages the best layer for document-level word alignment according to our layer search described in Appendix~\ref{app:hyp-selection}.

The gap between the final layer and layer 20 is large across all methods, aligning with prior observations~\citep{Zhang2019-fm, Jalili-Sabet2020-rb} as well as ours that the final layers of multilingual embedding models are often suboptimal for cross-lingual similarity tasks.

\section{Main Results with Precision, Recall and F1}
\label{app:full-results}

In Table~\ref{tab:alignment-results}, we report precision, recall, and F1 in addition to AER to complement the results shown in Figure~\ref{fig:test-results}. The non-AER metrics reveal a consistent precision--recall tradeoff between alignment strategies. Argmax-based variants favor precision and more often achieve the strongest F1 scores, whereas Itermax generally improves recall by recovering additional alignments at the cost of precision. This pattern holds across both \MDPAlign{} and \CTFAlign{}. Neither Itermax variant dominates recall consistently: \CTFAlign{}+Itermax achieves higher average recall with \texttt{LaBSE} and \texttt{Qwen}, while \MDPAlign{}+Itermax does so with \texttt{mmBERT}. Overall, the results show that the choice between Argmax and Itermax primarily controls the precision--recall balance, while the structural constraint determines how successfully this tradeoff is managed for a given encoder.

\begin{table*}[h]
\resizebox{\textwidth}{!}{
\begin{tabular}{l rrrr @{\hspace{8pt}} rrrr @{\hspace{8pt}} rrrr @{\hspace{8pt}} rrrr @{\hspace{8pt}} rrrr @{\hspace{8pt}} rrrr @{\hspace{8pt}}| rrrr}
\hline
& \multicolumn{4}{c}{\textbf{en-fr}} & \multicolumn{4}{c}{\textbf{en-ro}} & \multicolumn{4}{c}{\textbf{en-ja}} & \multicolumn{4}{c}{\textbf{en-zh}} & \multicolumn{4}{c}{\textbf{la-gr}} & \multicolumn{4}{c}{\textbf{en-cz}} & \multicolumn{4}{c}{\textbf{Avg}} \\
\textbf{Model} & \textbf{P} & \textbf{R} & \textbf{F1} & \textbf{AER} & \textbf{P} & \textbf{R} & \textbf{F1} & \textbf{AER} & \textbf{P} & \textbf{R} & \textbf{F1} & \textbf{AER} & \textbf{P} & \textbf{R} & \textbf{F1} & \textbf{AER} & \textbf{P} & \textbf{R} & \textbf{F1} & \textbf{AER} & \textbf{P} & \textbf{R} & \textbf{F1} & \textbf{AER} & \textbf{P} & \textbf{R} & \textbf{F1} & \textbf{AER} \\
\hline
\multicolumn{29}{l}{\textsc{Sentence-level}} \\
\textit{LaBSE} \SimAlign{} + Argmax & \underline{0.969} & \underline{0.948} & \underline{0.958} & \underline{0.040} & \underline{0.921} & \underline{0.676} & \underline{0.780} & \underline{0.220} & \underline{0.755} & 0.456 & \underline{0.569} & \underline{0.431} & \underline{0.880} & \underline{0.819} & \underline{0.848} & \underline{0.151} & 0.710 & 0.707 & 0.708 & 0.292 & \underline{0.883} & \underline{0.899} & \underline{0.891} & \underline{0.109} & \underline{0.853} & \underline{0.751} & \underline{0.792} & \underline{0.207} \\
\textit{mmBERT} \SimAlign{} + Argmax & 0.958 & 0.932 & 0.945 & 0.053 & 0.906 & 0.639 & 0.749 & 0.251 & 0.738 & 0.419 & 0.535 & 0.465 & 0.867 & 0.757 & 0.808 & 0.192 & 0.690 & 0.711 & 0.700 & 0.300 & 0.878 & 0.856 & 0.867 & 0.133 & 0.840 & 0.719 & 0.767 & 0.232 \\
\textit{Qwen} \SimAlign{} + Argmax & 0.932 & 0.826 & 0.876 & 0.117 & 0.795 & 0.527 & 0.634 & 0.366 & 0.647 & 0.354 & 0.458 & 0.542 & 0.843 & 0.670 & 0.747 & 0.253 & 0.646 & 0.774 & 0.704 & 0.297 & 0.809 & 0.796 & 0.802 & 0.197 & 0.779 & 0.658 & 0.704 & 0.295 \\
\textit{GPT-5.4-mini} 1-shot prompt & 0.831 & 0.787 & 0.808 & 0.185 & 0.552 & 0.574 & 0.563 & 0.437 & 0.270 & \underline{0.599} & 0.372 & 0.628 & 0.543 & 0.674 & 0.601 & 0.399 & \underline{0.788} & \underline{0.861} & \underline{0.823} & \underline{0.179} & 0.638 & 0.678 & 0.657 & 0.344 & 0.604 & 0.696 & 0.637 & 0.362 \\
\\
\hline
\multicolumn{29}{l}{\textsc{Document-level}} \\
\multicolumn{29}{l}{\hspace{1em}\textit{GPT-5.4-mini}} \\
1-shot prompt & 0.006 & 0.005 & 0.005 & 0.995 & 0.032 & 0.031 & 0.031 & 0.968 & 0.011 & 0.019 & 0.014 & 0.986 & 0.057 & 0.076 & 0.065 & 0.935 & 0.004 & 0.005 & 0.004 & 0.996 & 0.019 & 0.018 & 0.018 & 0.982 & 0.022 & 0.026 & 0.023 & 0.977 \\
\multicolumn{29}{l}{\hspace{1em}\textit{LaBSE}} \\
\SimAlign{} + Argmax & 0.940 & 0.859 & 0.898 & 0.097 & 0.905 & 0.596 & 0.719 & 0.281 & 0.630 & 0.253 & 0.361 & 0.639 & 0.842 & 0.715 & 0.773 & 0.227 & 0.446 & 0.308 & 0.364 & 0.635 & 0.873 & 0.791 & 0.830 & 0.168 & 0.773 & 0.587 & 0.657 & 0.341 \\
\MDPAlign{} + Argmax (fuzzy) & 0.938 & 0.847 & 0.890 & 0.103 & \textbf{0.926} & 0.658 & 0.769 & 0.231 & 0.567 & 0.275 & 0.370 & 0.630 & 0.857 & 0.800 & \textbf{0.828} & \textbf{0.172} & \textbf{0.653} & 0.598 & 0.624 & 0.375 & 0.877 & 0.868 & 0.872 & 0.127 & 0.803 & 0.674 & 0.725 & 0.273 \\
\MDPAlign{} + Itermax (fuzzy) & 0.887 & 0.894 & 0.890 & 0.110 & 0.809 & \textbf{0.748} & \textbf{0.777} & \textbf{0.222} & 0.477 & 0.350 & 0.404 & 0.596 & 0.747 & \textbf{0.882} & 0.809 & 0.191 & 0.533 & 0.680 & 0.598 & 0.405 & 0.777 & \textbf{0.931} & 0.847 & 0.158 & 0.705 & 0.748 & 0.721 & 0.280 \\
\CTFAlign{} + Argmax & \textbf{0.958} & 0.899 & \textbf{0.928} & \textbf{0.068} & 0.921 & 0.653 & 0.764 & 0.235 & \textbf{0.706} & 0.378 & 0.492 & 0.507 & 0.854 & 0.791 & 0.821 & 0.179 & 0.651 & 0.601 & 0.625 & 0.375 & \textbf{0.879} & 0.869 & \textbf{0.874} & \textbf{0.126} & \textbf{0.828} & 0.699 & \textbf{0.751} & \textbf{0.248} \\
\CTFAlign{} + Itermax & 0.904 & \textbf{0.934} & 0.919 & 0.084 & 0.807 & 0.743 & 0.774 & 0.226 & 0.597 & \textbf{0.459} & \textbf{0.519} & \textbf{0.481} & 0.743 & 0.869 & 0.801 & 0.199 & 0.522 & 0.658 & 0.582 & 0.419 & 0.779 & 0.927 & 0.847 & 0.159 & 0.725 & \textbf{0.765} & 0.740 & 0.261 \\
\multicolumn{29}{l}{\hspace{1em}\textit{mmBERT}} \\
\SimAlign{} + Argmax & 0.742 & 0.418 & 0.535 & 0.457 & 0.866 & 0.419 & 0.565 & 0.435 & 0.566 & 0.115 & 0.191 & 0.809 & 0.854 & 0.598 & 0.703 & 0.296 & 0.274 & 0.125 & 0.172 & 0.828 & 0.811 & 0.474 & 0.598 & 0.400 & 0.686 & 0.358 & 0.461 & 0.537 \\
\MDPAlign{} + Argmax (fuzzy) & 0.803 & 0.684 & 0.739 & 0.254 & \textbf{0.890} & 0.563 & 0.690 & 0.310 & 0.404 & 0.166 & 0.235 & 0.764 & \textbf{0.863} & 0.745 & \textbf{0.800} & \textbf{0.200} & 0.458 & 0.448 & 0.453 & 0.547 & 0.819 & 0.698 & 0.754 & 0.245 & 0.706 & 0.551 & 0.612 & 0.387 \\
\MDPAlign{} + Itermax (fuzzy) & 0.670 & 0.779 & 0.720 & 0.288 & 0.771 & \textbf{0.646} & \textbf{0.703} & \textbf{0.297} & 0.328 & 0.214 & 0.259 & 0.741 & 0.736 & \textbf{0.849} & 0.788 & 0.212 & 0.372 & 0.553 & 0.445 & 0.558 & 0.693 & 0.804 & 0.744 & 0.258 & 0.595 & 0.641 & 0.610 & 0.392 \\
\CTFAlign{} + Argmax & 0.758 & 0.576 & 0.655 & 0.332 & 0.878 & 0.549 & 0.676 & 0.325 & 0.542 & 0.216 & 0.309 & 0.691 & 0.847 & 0.711 & 0.773 & 0.227 & 0.226 & 0.173 & 0.196 & 0.803 & \textbf{0.830} & 0.693 & 0.755 & 0.243 & 0.680 & 0.486 & 0.561 & 0.437 \\
\CTFAlign{} + Itermax & 0.513 & 0.451 & 0.480 & 0.515 & 0.754 & 0.605 & 0.671 & 0.329 & 0.462 & 0.292 & 0.358 & 0.642 & 0.729 & 0.810 & 0.767 & 0.233 & 0.275 & 0.359 & 0.311 & 0.690 & 0.707 & 0.796 & 0.749 & 0.252 & 0.573 & 0.552 & 0.556 & 0.444 \\
\multicolumn{29}{l}{\hspace{1em}\textit{Qwen}} \\
\SimAlign{} + Argmax & 0.856 & 0.644 & 0.735 & 0.252 & 0.837 & 0.510 & 0.634 & 0.366 & \textbf{0.631} & 0.291 & 0.398 & 0.602 & 0.802 & 0.649 & 0.717 & 0.282 & 0.564 & 0.550 & 0.557 & 0.443 & 0.828 & 0.757 & 0.791 & 0.208 & 0.753 & 0.567 & 0.639 & 0.359 \\
\MDPAlign{} + Argmax (fuzzy) & 0.837 & 0.691 & 0.757 & 0.232 & 0.832 & 0.545 & 0.659 & 0.342 & 0.401 & 0.180 & 0.248 & 0.752 & 0.788 & 0.708 & 0.746 & 0.254 & 0.556 & 0.662 & 0.604 & 0.397 & 0.806 & 0.752 & 0.778 & 0.221 & 0.703 & 0.590 & 0.632 & 0.366 \\
\MDPAlign{} + Itermax (fuzzy) & 0.750 & 0.777 & 0.763 & 0.239 & 0.715 & 0.631 & 0.670 & 0.330 & 0.330 & 0.231 & 0.272 & 0.728 & 0.662 & 0.811 & 0.729 & 0.271 & 0.436 & 0.739 & 0.548 & 0.455 & 0.672 & 0.839 & 0.746 & 0.257 & 0.594 & 0.671 & 0.621 & 0.380 \\
\CTFAlign{} + Argmax & \textbf{0.895} & 0.792 & \textbf{0.840} & \textbf{0.151} & 0.833 & 0.548 & 0.661 & 0.339 & 0.629 & 0.369 & 0.465 & 0.535 & 0.789 & 0.706 & 0.745 & 0.255 & \textbf{0.643} & 0.790 & \textbf{0.709} & \textbf{0.293} & 0.810 & 0.809 & \textbf{0.809} & \textbf{0.191} & \textbf{0.767} & 0.669 & \textbf{0.705} & \textbf{0.294} \\
\CTFAlign{} + Itermax & 0.792 & \textbf{0.863} & 0.826 & 0.181 & 0.711 & 0.631 & 0.669 & 0.332 & 0.519 & \textbf{0.449} & \textbf{0.481} & \textbf{0.518} & 0.660 & 0.802 & 0.724 & 0.276 & 0.505 & \textbf{0.845} & 0.632 & 0.371 & 0.685 & \textbf{0.881} & 0.771 & 0.233 & 0.645 & \textbf{0.745} & 0.684 & 0.319 \\
\hline
\end{tabular}
}
\caption{Detailed word alignment test set results (P, R, F1, AER) across language pairs and models. Sentence-level rows are upper-bound baselines; best per column \underline{underlined}. Document-level best per column \textbf{bold}.}
\label{tab:alignment-results}
\end{table*}

\section{\texttt{mmBERT} Failure Case}
\label{app:fail}

\begin{figure}[h]
    \centering

    \begin{minipage}{0.48\columnwidth}
        \centering
        \includegraphics[width=\linewidth]{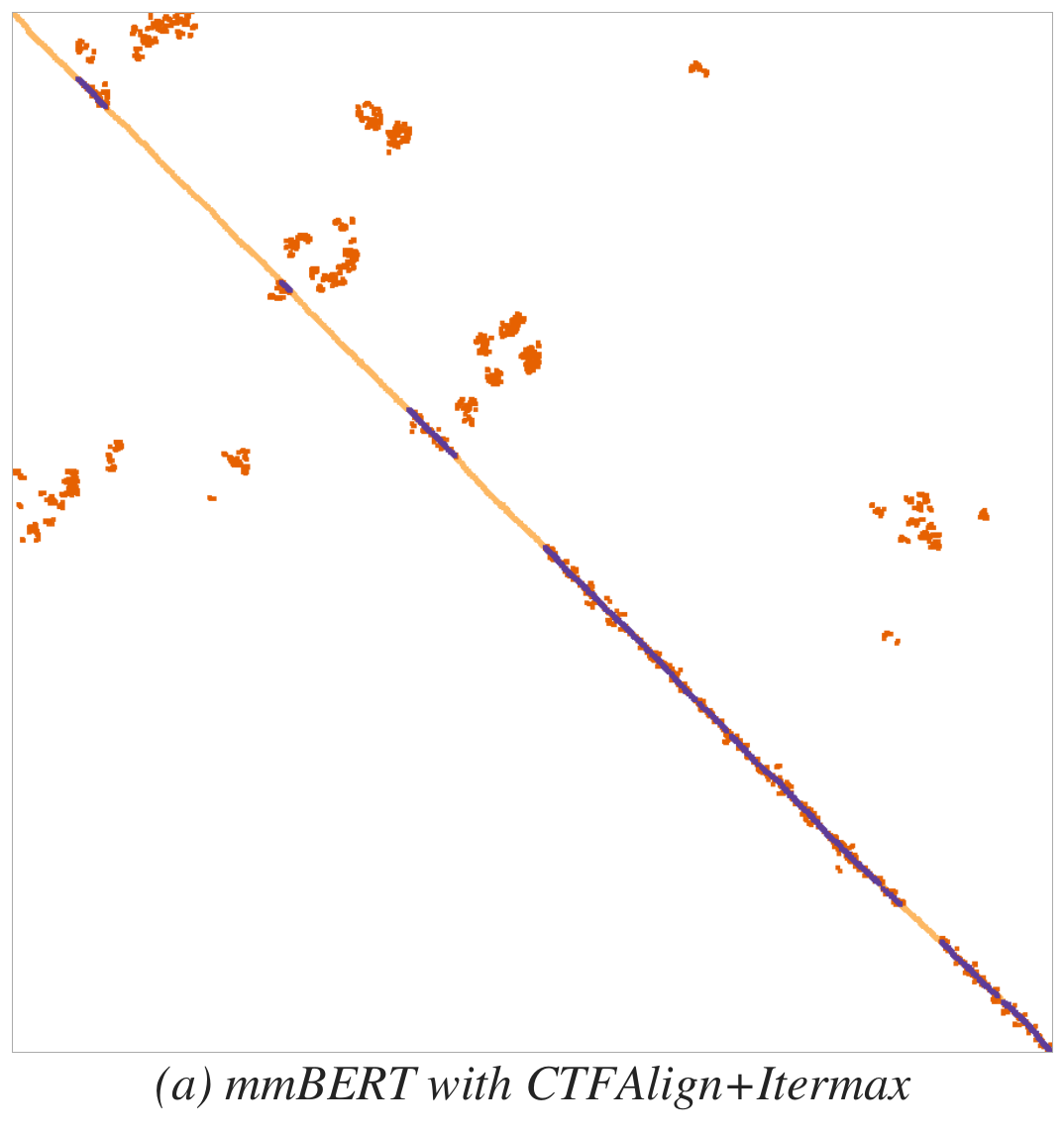}
    \end{minipage}
        \hfill
    \begin{minipage}{0.48\columnwidth}
        \centering
        \includegraphics[width=\linewidth]{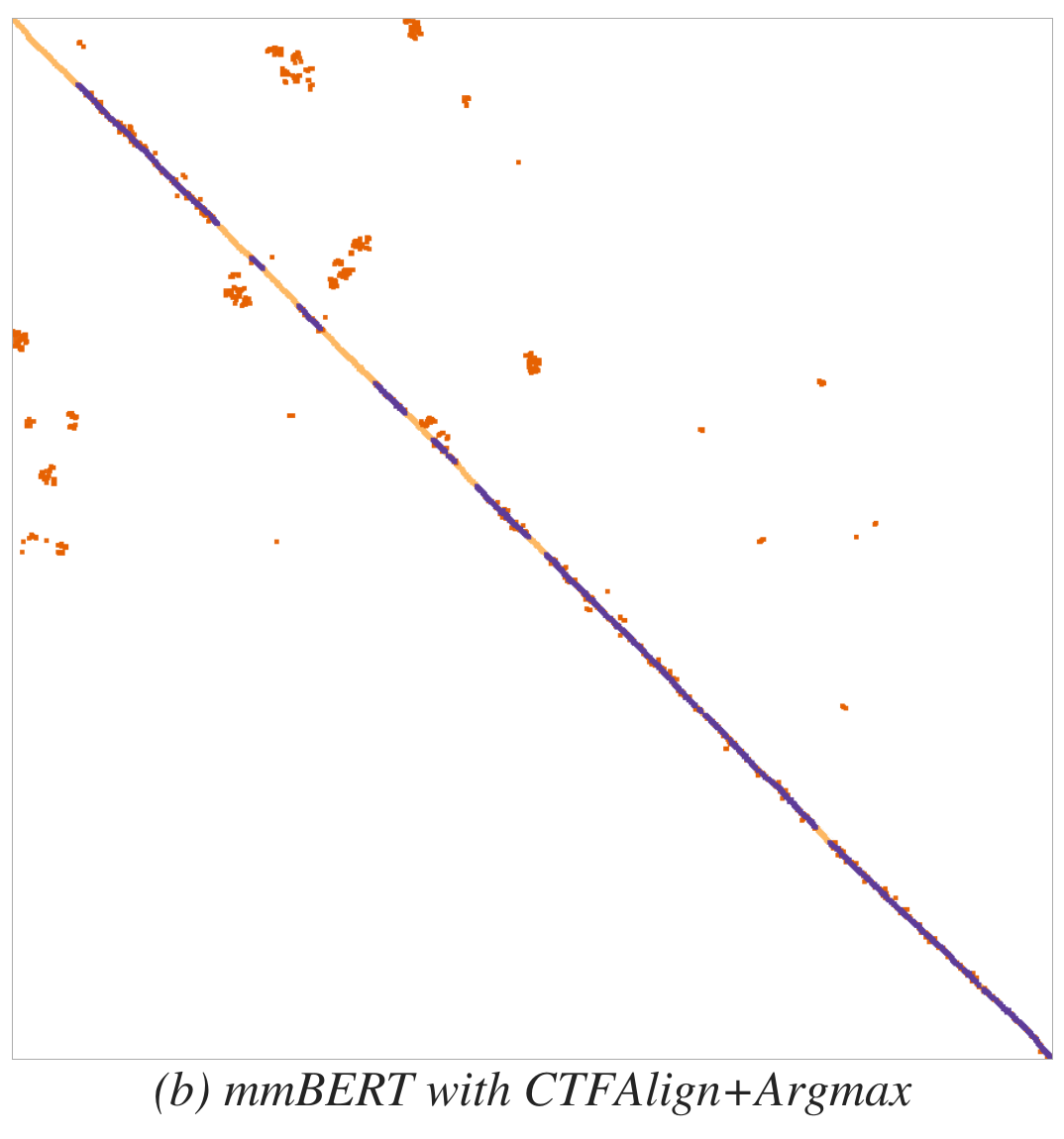}
    \end{minipage}
    
    \caption{(a) \CTFAlign{} + Itermax with \texttt{mmBERT} failure cases on en--fr in comparison to partially recovered alternative (b) \CTFAlign{} + Argmax. Colors denote \textcolor{mypurple}{correct alignments}, \textcolor{mylightorange}{false negatives}, \textcolor{myorange}{false positives}.}
    \label{fig:fail}
\end{figure}

In Figure~\ref{fig:fail}, we visualize the failure case on the en--fr test set example with \texttt{mmBERT} embeddings and \CTFAlign{}+Itermax. The failure case illustrates that when combining \CTFAlign{} with Itermax on a noisy similarity matrix the approach is too lenient and results in spurious alignment regions, which are likely the result of falsely aligned coarse similarity blocks at a lower resolution.

\end{document}